\documentclass[pre-print,3p,times,fleqn]{elsarticle}
\usepackage{amsmath}
\usepackage{amssymb}
\usepackage{psfrag}
\usepackage{color}
\usepackage{stfloats}
\usepackage{fixltx2e}
\usepackage{mathrsfs}
\usepackage{tabularx}
\usepackage{booktabs}
\usepackage{colortbl}
\usepackage{rotating}
\usepackage{boldline}
\usepackage{url} 
\usepackage{bm}
\usepackage{changes}
\usepackage{graphicx}
\usepackage{caption,subcaption}
\usepackage{comment}
\usepackage{pgf}
\usepackage{diagbox}
\usepackage{tikz}
\usepackage{extpfeil}
\usepackage{relsize}
\usepackage{lineno}
\usetikzlibrary{arrows,automata}

\newcommand\com[1]{\color{black}#1}

\journal{Computer Methods in Applied Mechanics and Engineering}
\begin{document}
\renewcommand{\topfraction}{0.98}	
\renewcommand{\bottomfraction}{0.98}
\setcounter{topnumber}{3}
\setcounter{bottomnumber}{3}
\setcounter{totalnumber}{4}         
\setcounter{dbltopnumber}{4}        
\renewcommand{\dbltopfraction}{0.98}	
\renewcommand{\textfraction}{0.05}	
\renewcommand{\floatpagefraction}{0.5}	
\renewcommand{\dblfloatpagefraction}{0.5}	
\newcommand{\beq}{\begin{equation}}
\newcommand{\eeq}{\end{equation}}
\newcommand{\divg}{\mbox{\rm{div}}\,}
\newcommand{\Divg}{\mbox{\rm{Div}}\,}
\newcommand{\D}  {\displaystyle}
\newcommand{\DS} {\displaystyle}
\newcommand{\RM}[1]{\textit{\MakeUppercase{\romannumeral #1{}}}}
\newtheorem{remark}{\bf{{Remark}}}
\def\sca   #1{\mbox{\rm{#1}}{}}
\def\mat   #1{\mbox{\bf #1}{}}
\def\vec   #1{\mbox{\boldmath $#1$}{}}
\def\scas  #1{\mbox{{\scriptsize{${\rm{#1}}$}}}{}}
\def\scaf  #1{\mbox{{\tiny{${\rm{#1}}$}}}{}}
\def\vecs  #1{\mbox{\boldmath{\scriptsize{$#1$}}}{}}
\def\tens  #1{\mbox{\boldmath{\scriptsize{$#1$}}}{}}
\def\tenf  #1{\mbox{{\sffamily{\bfseries {#1}}}}}
\def\ten   #1{\mbox{\boldmath $#1$}{}}
\def\Ass  {\overset{\hspace*{0.4cm} n_{\scas{el}}}
          {\underset{\scaf{c},\scaf{d}=1}{\msf{A}}}}
\def\ltr   #1{\mbox{\sf{#1}}}
\def\bltr  #1{\mbox{\sffamily{\bfseries{{#1}}}}}
\sloppy
\begin{frontmatter}
\title{\Large Machine learning in cardiovascular flows modeling:\\
{\com Predicting arterial blood pressure from non-invasive 4D flow MRI data using physics-informed neural networks}} 
\author[1]{Georgios Kissas}
\author[1]{Yibo Yang}
\author[2]{Eileen Hwuang}
\author[3]{Walter R. Witschey}
\author[4]{John A. Detre}
\author[1]{Paris Perdikaris$^*$}

\ead{pgp@seas.upenn.edu, corresponding author}

\address[1]{Department of Mechanical Engineering and Applied Mechanics, University of Pennsylvania, Philadelphia, Pennsylvania, USA}

\address[2]{Department of Bioengineering, University of Pennsylvania, Philadelphia, Pennsylvania, USA}

\address[3]{Department of Radiology, Perelman School of Medicine, University of Pennsylvania, Philadelphia, Pennsylvania, USA}

\address[4]{Department of Neurology, Perelman School of Medicine, University of Pennsylvania, Philadelphia, Pennsylvania, USA}

%
\begin{abstract} %
Advances in computational science offer a principled pipeline for predictive modeling of cardiovascular flows and aspire to provide a valuable tool for monitoring, diagnostics and surgical planning. Such models can be nowadays deployed on large patient-specific topologies of systemic arterial networks and return detailed predictions on flow patterns, wall shear stresses, and pulse wave propagation. However, their success heavily relies on tedious  pre-processing and calibration procedures that typically induce a significant computational cost, thus hampering their clinical applicability. In this work we put forth a machine learning framework that enables the seamless synthesis of non-invasive {\it in-vivo} measurement techniques and computational flow dynamics models derived from first physical principles. We illustrate this new paradigm by showing how one-dimensional models of pulsatile flow can be used to constrain the output of deep neural networks such that their predictions satisfy the conservation of mass and momentum principles. Once trained on noisy and scattered clinical data of flow and wall displacement, these networks can return physically consistent predictions for velocity, pressure and wall displacement  pulse wave propagation, all without the need to employ conventional simulators. A simple post-processing of these outputs can also provide a relatively cheap and effective way for estimating Windkessel model parameters that are required for the calibration of traditional computational models. The effectiveness of the proposed techniques is demonstrated through a series of prototype benchmarks, as well as a realistic clinical case involving {\it in-vivo} measurements near the aorta/carotid bifurcation of a healthy human subject.
\end{abstract}
\end{frontmatter}


\section{Introduction}
\label{sec:introdunction}


Recent advances in clinical measurement and computational modeling techniques introduce new capabilities for monitoring the human cardiovascular system from different perspectives such as disease surveys \cite{rose1982cardiovascular}, bio-medical image processing \cite{kak2002principles,haris2014technique}, computational mathematics \cite{formaggia2010cardiovascular, figueroa2006coupled, reymond2009validation,grinberg2011modeling}, bio-physics \cite{stefanovska1999physics, ma2018relation}, etc. These studies reveal the crucial role played by blood flow, arterial wall mechanics and pressure wave propagation, and how their interplay directly characterizes the functionality of the cardiovascular system both in health and disease (e.g., hypertensive disorders \cite{o1995mechanical}).


Understanding the inner workings of the cardiovascular system has been central to many studies involving clinical, interventional or computational approaches. For instance, Chan {\it et. al.} \cite{chan2007hybrid} proposed placing sensors in the human body  to achieve real time measuring of arterial velocity and wall displacement to monitor the health of patients. Although the collected {\it in-vivo} measurements can be highly accurate, such interventional techniques are sometimes expensive and suffer from limitations that are not easy to address, e.g., difficulties of placing probes in cerebral or uteroplacental arteries \cite{kett2002adverse}. These limitations motivate the use of non-invasive measurement techniques such as bio-medical imaging, advances in which currently define the clinical standard of care \cite{edelstein1980spin, antiga2008image, yushkevich2006user, SCI:Seg3D}. To this end, one of the most commonly used techniques is Doppler ultrasound velocimetry \cite{aaslid1982noninvasive} that enables the recovery of blood velocity wave propagation. Using information related to the  pulsatility of the flow \cite{mitchell2011arterial}, clinicians are able to infer quantitative information about the pressure in the arterial vessels, e.g., large pulsatility is caused by large pressure \cite{ma2018relation}. Rather than Doppler ultrasound, more recent studies leverage advances in 4D flow Magnetic Resonance Imaging (MRI) \cite{markl20124d} to recover the full three-dimensional velocity flow field. This method provides a more detailed representation of the flow, and also a larger spatial coverage that can capture quantities like vessel tortuosity in a more accurate manner. Moreover, structural characteristics like the cross-sectional area and vessel diameter can be recovered by segmenting 2D cine images \cite{plein2001steady} from MRI measurements. However, critical variables such as the pressure cannot be directly measured by a non-invasive technique \cite{barker1985non}, and accurate measurements are only accessible by inserting a catheter equipped with sensors inside the vessel of interest. 


These difficulties of directly measuring quantities of interest like the pressure in an {\it in-vivo} and non-invasive manner have motivated the use of computer simulations and computational fluid dynamics models to predict them {\it in-silico}. Advances in algorithms and computing resources now allow us to perform detailed flow simulations in complex patient-specific arterial topologies using three-dimensional (3D) and/or one-dimensional (1D) formulations of the unsteady Navier-Stokes equations \cite{formaggia2010cardiovascular, figueroa2006coupled, reymond2009validation,grinberg2011modeling, urquiza2006multidimensional, sherwin2003one}. Such tools have been successfully validated against both {\it in-vitro} experiments \cite{matthys2007pulse} as well as {\it in-vivo} clinical data \cite{revie2013validation}, and provide a valuable platform for parametric sensitivity studies \cite{chen2013simulation}. Despite their predictive power, computational models have still not made their way into clinical practice primarily due to their high computational cost and the tedious procedures needed for their practical deployment, e.g., mesh generation, parameter calibration, etc. For instance, such models require the precise subscription of boundary conditions that effectively capture the downstream flow dynamics in small arteries and arterioles via the use of Windkessel or structured tree models \cite{westerhof2009,grinberg2008outflow,olufsen1999structured,perdikaris2015effective}. Inaccurate calibration of the parameters associated with these boundary conditions is often the cause of brittleness in the resulting predictions, thus limiting the translational impact of computational models to the clinical domain.

This goal of this work is to put forth a new paradigm for seamlessly blending non-invasive measurement techniques such 4D flow MRI with computational fluid dynamics models, towards providing a cheap and effective tool for characterizing velocity and pressure pulse wave propagation in human systemic arteries. Leveraging recent advances in physics-informed machine learning \cite{raissi2019physics, raissi2017machine, yang2018adversarial, zhu2018bayesian}, we put forth an algorithmic framework for producing physically consistent predictions of flow and pressure wave propagation directly from processing noisy and scattered measurements of blood velocity and wall displacement obtained by non-invasive 4D flow MRI. Specifically, we propose to employ deep neural networks to represent the unknown flow variables (blood velocity, wall displacement and pressure) in a given arterial network. We then train these networks to produce outputs that: (i) fit any available clinical data (typically velocity and wall displacement data at a few cross-sections), (ii) satisfy the underlying physical  conservation laws as described by a reduced one-dimensional model of pulsatile blood flow \cite{sherwin2003one,formaggia2010cardiovascular}, and (iii) ensure conservation and information propagation across interface points in the network (e.g., bifurcations, junctions, etc.). 

In contrast to traditional computational fluid dynamics models, the proposed methodology does not the cumbersome generation of computational meshes, nor it requires the precise prescription of boundary conditions. Instead, it can directly use noisy and scattered measurements obtained by medical imaging at points other than the boundaries to accurately predict the pressure and recover flow information. Furthermore, as flow conditions in general do not vary greatly from one patient to another, pre-trained neural network models can be rapidly adapted to a new patient case. After the model is trained, all quantities of interest (blood velocity, wall displacement and pressure) can be predicted for every temporal or spatial point in the network. This directly enables the calculation of Windkessel parameters as a simple post-processing step, leading to a simple scheme for calibrating more detailed and expensive 3D models of blood flow.

{\com
The main contributions of this work can be summarized in the following bullet points:
\begin{itemize}
    \item This is the first time that physics-informed neural networks are used to solve conservation laws in graph topologies (specifically, here we considered disjoint arterial domains connected by appropriate interface boundary conditions).
    \item In this paper the importance of non-dimensionalization and normalization is highlighted and a methodology to properly mitigate vanishing gradient pathologies in physics-informed neural networks. 
    \item To our knowledge, this is the first time that a physics-informed neural network model is applied on an inference problem involving real noisy clinical data.
    \item We were able to reconstruct the spatio-temporal response of the absolute pressure wave in an arterial network, given scattered and noisy 4D flow MRI data of blood velocity velocity and wall displacement at some cross-sections.
    \item This contribution is a first step towards utilizing physics-informed deep learning for performing non-invasive pressure diagnostics for cardiovascular disorders.
    \item The resulting surrogate also offers a simple and computationally efficient way for calibrating boundary conditions for conventional flow simulation models.
\end{itemize}
}

The rest of the paper is organized as follows. In section (\ref{sec:methods}), we provide an outline of the one-dimensional blood flow model employed in this work, as well as the deep learning techniques used for obtaining physically consistent predictions of pulse wave propagation directly from clinical measurements. In section (\ref{sec:results}), we demonstrate the effectiveness of the proposed methods in a series of synthetic systematic studies, as well as a realistic clinical case involving {\it in-vivo} measurements near the aorta/carotid bifurcation of a healthy human subject. Conclusions and further discussions are given in section (\ref{sec:discussion}). All code and data accompanying this manuscript will be made publicly available at \url{https://github.com/PredictiveIntelligenceLab/1DBloodFlowPINNs}.

\section{Methods}
\label{sec:methods}


\subsection{A simplified model of pulsatile flow in arterial networks}
\label{sec:simplified_model}
\begin{figure}
\centering
\includegraphics[width=0.8\textwidth]{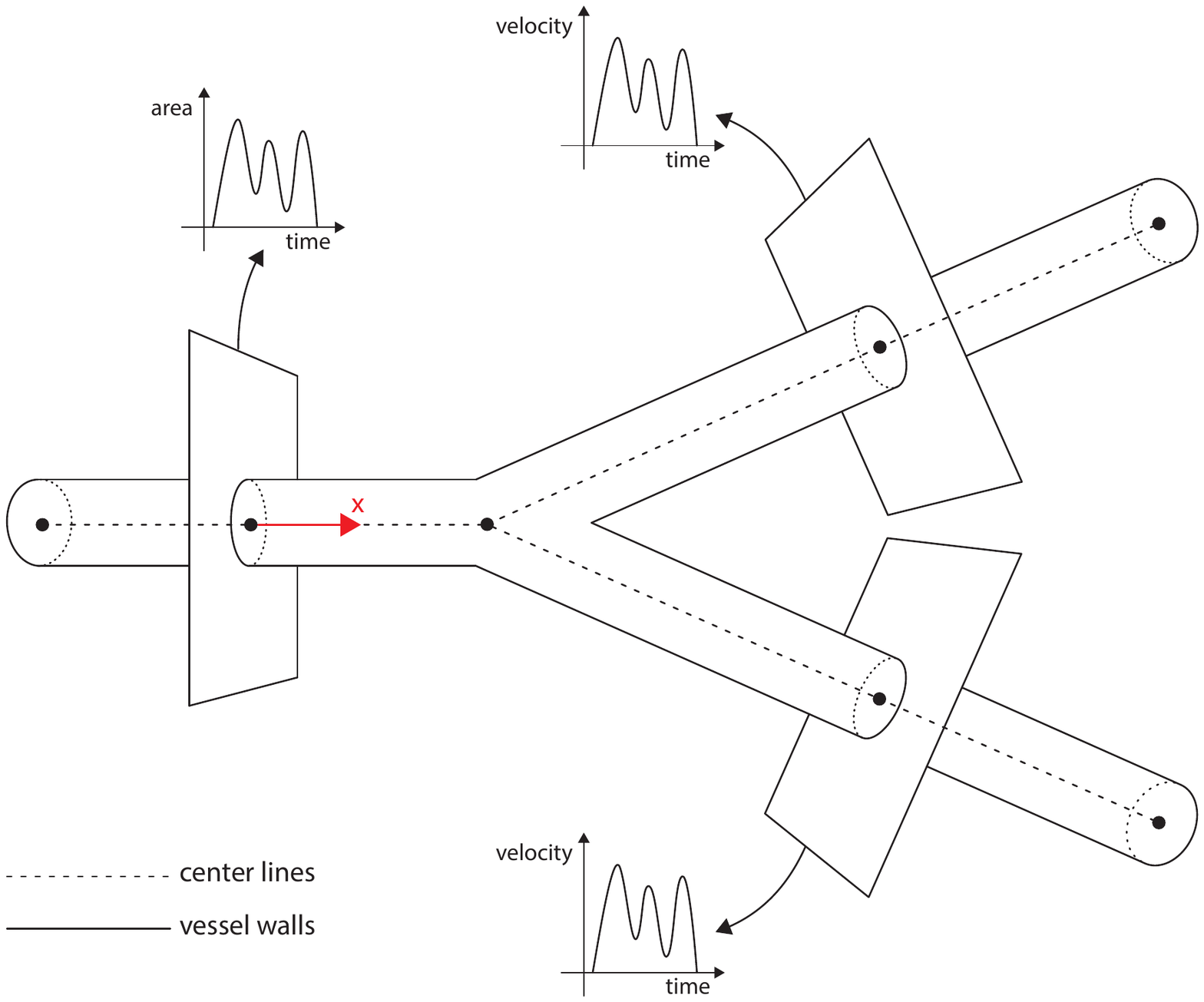}
\caption{{\em A schematic representation of a simple bifurcating arterial topology and its one-dimensional center-lines used for the reduced model.} Throughout this work, three dimensional geometries are recovered from MRI data and the corresponding center-lines are extracted by using the VMTK library  \cite{antiga2008image}. Blood velocity data are typically obtained using Doppler ultrasound or 4D flow MRI, while the area data are parsed from 2D Cine images recovered by 4D flow MRI. }
\label{fig:1dgeometry}
\end{figure}

Pulse wave propagation in arterial networks can be effectively modeled using one-dimensional (1D) reduced order models \cite{sherwin2003one, reymond2009validation, reymond2011validationPS}. In order to achieve the order reduction, a series of assumptions need to be made  \cite{sherwin2003computational}. First, we assume that the local curvature is small enough such that the geometry can be described using a Cartesian coordinate $x$, as shown in figure \ref{fig:1dgeometry}. Moreover, the fluid is incompressible and Newtonian, since we are considering geometries consisting of large arteries, so the density and dynamic viscosity are constant. Lastly, the structural properties of the artery are preserved at a cross-section. Following \cite{sherwin2003computational, formaggia2010cardiovascular} we consider a reduced form of the incompressible Navier-Stokes equations. Conservation of mass and momentum can be expressed as a hyperbolic conservation law \cite{sherwin2003computational} that describes the evolution of blood velocity and cross-sectional area \cite{sherwin2003one, formaggia2010cardiovascular}. In order to close the system, a third equation accounting for the relation between the pressure and the area is used, which is derived by assuming a thin wall tube and using Laplace's law \cite{sherwin2003computational}. 
This reduced order model provides an accurate representation of the underlying transport processes and its effectiveness in correctly capturing pulse wave propagation phenomena has been validated against both {\it in-vitro} and {\it in-vivo} data \cite{matthys2007pulse,reymond2009validation, reymond2011validationPS}. 
The system derived by the above analysis takes the form \cite{sherwin2003one, formaggia2010cardiovascular}

\begin{equation}
\begin{split} 
       & \frac{\partial A}{\partial t} + \frac{\partial A u }{\partial x} = 0, \\
    & \frac{\partial u }{\partial t} + \alpha u \frac{\partial u}{\partial x} + \frac{u}{A} \frac{\partial (\alpha -1)uA}{\partial x} + \frac{1}{\rho} \frac{\partial p }{\partial x} - K_R \frac{u}{A}= 0, \\
    & p = p_{ext} + \beta ( \sqrt{A} - \sqrt{A_0}),
\end{split}
\label{equ:systemOfEquations}
\end{equation}
where  $A(x,t)$, $u(x,t)$ and $p(x,t)$ represent the cross-sectional area, the velocity and pressure, respectively. In addition, $x$ and $t$ represent the spatial and temporal coordinate in each vessel, respectively. Also, $\beta = \dfrac{\sqrt{\pi} h_0 E}{(1 - \nu^2)A_0}$ and $A_0$ denotes the vessel's sectional area at equilibrium, $h_0$ the wall thickness, $E$ the Young's modulus, $p_{ext}$ the external pressure, $\nu$ the Poisson ratio and $\rho$ is the density of the blood. Values for $h_0$, $E$ and $\nu$ are typically mined from the literature, while $p_{ext}$ and $A_0$ are typically set to the diastolic pressure and wall displacement, respectively. {\color{black} Moreover, $K_{R}$ is a friction parameter that depends on the velocity profile chosen, while $\alpha$ is a momentum flux correction factor which accounts for the nonlinearity of the sectional integration in terms of the local velocity (for blood flow we typically chose $K_{R} = -22\mu \pi$, where $\mu$ is the blood viscosity, and $\alpha=1.1$, see \cite{sherwin2003one} for more details)}. For the examples presented in section (\ref{sec:results}) the $\beta$ parameter is computed based on the empirical relation put forth by Olufsen {\it et. al.} \cite{olufsen1999structured}. {\color{black} We should also note that the proposed one-dimensional model neglects vessel tortuosity, curvature and other three-dimensional geometry effects. In cases where these play a major role, e.g., cerebral or coronary flows, our model in its present form will be inevitably prone to inaccuracies. Several works in the literature have attempted to extend one-dimensional models to account for three-dimensional geometry effects (see for e.g., \cite{formaggia2010cardiovascular}, \cite{lamponi2004one}). Such extensions could be incorporated in our framework and are left to future work as they extend beyond the scope of this paper. For all cases presented in this work, we reckon that vessel tortuosity has a secondary effect in comparison with other sources of inaccuracy/uncertainty (e.g., noise in the experimental data, uncertainty on the geometry, the wall elasticity moduli, etc.).}

\subsection{Physics-informed neural networks}
\begin{figure}
\centering
\includegraphics[width=\textwidth]{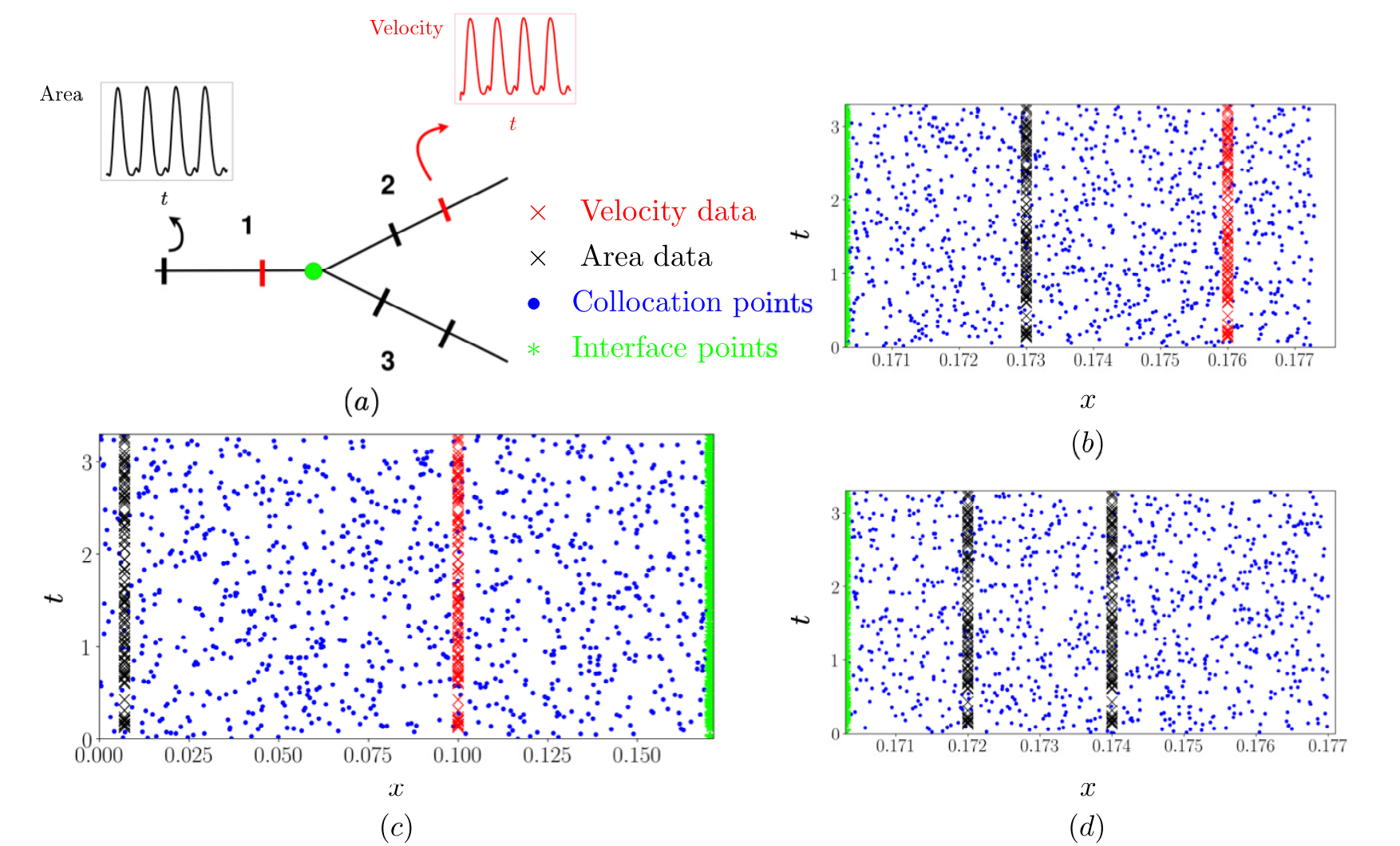}
\caption{{\em Graph of a prototype arterial network where 1 vessel splits into 2 vessels at the bifurcation point:} (a) Structure of the simple arterial network consisting of 3 vessels with 1 bifurcation point. (b) Domain of vessel $\#2$. (c) Domain of vessel $\#1$. (d) Domain of vessel $\#3$. Blue points represent the collocation points. Green stars represent the interface points at the bifurcation. Red crosses indicate the locations of the velocity measurements. Black crosses show the locations for the area measurements.}
\label{fig:domain_1bifurcation}
\end{figure}

Physics informed neural networks (PINNs) is a new paradigm  that leverages recent advances in deep learning to infer solutions, parameters and/or constitutive laws involving partial differential equations (PDEs) \cite{raissi2017physics,raissi2017physic, raissi2019physics, tartakovsky2018learning}. In this framework, the solution of partial differential equations is parametrized by a neural network that is trained to match the measurements of the system, while being constrained to approximately satisfy the underlying physical  laws. In our particular case, we define one neural network $f(x,t; \theta^j)$ to represent the solution of the equation (\ref{equ:systemOfEquations}) for vessel $\# j$ in our arterial network. Here $\theta^j$ denotes the parameters of the neural network for vessel $\# j$. The network aims to approximate the following mapping:
$$\mathlarger{[x, t]\xmapsto[ ]{f_{\theta^j}}[A^j(x, t), u^j(x, t), p^j(x, t)]}$$
In correspondence to equation (\ref{equ:systemOfEquations}), the following residuals can be defined:
\begin{equation}
\begin{split} 
       & r_{A}(x,t) :=  \frac{\partial A}{\partial t} + \frac{\partial A u }{\partial x}, \\
    & r_{u}(x,t) :=  \frac{\partial u }{\partial t} + \alpha u \frac{\partial u}{\partial x} + \frac{u}{A} \frac{\partial (\alpha -1)uA}{\partial x} + \frac{1}{\rho} \frac{\partial p }{\partial x} - K_R \frac{u}{A},\\
    & r_{p}(x,t) :=  p - p_{ext} - \beta ( \sqrt{A} - \sqrt{A_0}).
\end{split}
\label{equ:PINNS_residual}
\end{equation}
These residuals can be then used as constrains during the training of the neural networks $f(x,t; \theta^j)$ in order to encourage them to produce physically consistent predictions. In addition to minimizing the residuals, the neural networks are also constrained to fit any available non-invasive clinical measurements that may be noisy and scattered in space and time. In order to illustrate our model, we refer figure (\ref{fig:domain_1bifurcation}). For example, in each vessel $\# j$, we have scattered measurements of $A(x,t)$ (corresponding to the black $\times$'s in figure (\ref{fig:domain_1bifurcation})), namely $\{(x_{i}, t_i), A^j(x_{i}, t_i)\}$, $i = 1,\dots,N_A^j$, and $u(x,t)$ (corresponding to the red $\times$'s in figure (\ref{fig:domain_1bifurcation})), namely $\{(x_{i}, t_i), u^j(x_{i}, t_i)\}$, $i = 1,\dots,N_u^j$, along with a larger number of collocation points $\{(x_{i}, t_i), r_{A,u,p}^j(x_{i}, t_i) = 0\}$, $i = 1,\dots,N_r^j$, that aim to satisfy the constraints at a finite set of $N_r^j$ collocation nodes. In the examples presented in this paper, we want to emphasize that we will only use measurements for $A(x,t)$ and $u(x,t)$, but not for $p(x,t)$, as these are the quantities that often cannot be reliably measured non-invasively in the clinic. 

{\color{black} The formulation of the one-dimensional model employed in this work introduces an explicit correlation between the absolute pressure, the cross-sectional area and the blood velocity, as shown in equation (\ref{equ:systemOfEquations}). Utilizing momentum conservation alone as a constraint is not sufficient for identifying the absolute pressure as only the pressure gradient appears in the momentum equation. However, the linear elastic constitutive law for the wall displacements (see equation (\ref{equ:systemOfEquations})) directly relates the arterial wall displacement with the absolute pressure in each cross section.
This constitutive relation is coupled to the mass and momentum conservation laws, and it is exactly these correlations that we aim to exploit via the use of physics-informed neural networks in order to infer the absolute pressure from velocity and cross-sectional area measurements. 
Of course the sensible choice of sensible values for $p_{ext}$ and $A_0$ plays a role in the convergence speed of the proposed algorithm, but this choice is not unique. In the synthetic examples presented in this work $p_{ext}$ was set to zero and $A_0$ was set to a reference value at equilibrium. For the realistic aorta/carotid bifurcation case the values of $p_{ext}$ and $A_0$ in each vessel were chosen according to nominal values for the diastolic pressure and area found in the literature (see e.g. \cite{xiao2014systematic}).} More details on how to construct the respective terms of the loss function for training the neural networks $f(x,t; \theta^j)$ will be given in the next section.

\subsection{Loss function: Measurements}\label{sec:Loss_measurement}
This part of the loss function corresponds to fitting the clinical measurements obtained for some of the vessels in the network. In figure (\ref{fig:domain_1bifurcation}) these are denoted as black and red $\times$'s, the color depending on the quantity we have measurement for. 
In this particular case, this term encourages the output of the neural network $A$ and $u$ to match the measurements of area and velocity  obtained by a clinical procedure (e.g., segmenting 2D cine images and Doppler ultrasound \cite{aaslid1982noninvasive}). This part of the loss function has the form (take vessel $\# j$ as example):
\begin{align}
    \mathcal{L}_{\textrm{measurements}}^j = \frac{1}{N_u^j} \sum_{i=1}^{N_u^j} ( u^j (x_i, t_i) - u^j(x_i, t_i; \theta^j))^2 + \frac{1}{N_A^j}\sum_{i=1}^{N_A^j} ( A^j (x_i, t_i) - A^j(x_i, t_i; \theta^j))^2, \quad \quad j = 1, \dots, D_M
\end{align}
where $N_A^j$, $N_u^j$ represent the number of measurements for $A$ and $u$ in vessel $\# j$ respectively. Also, $D_M$ denote total number of vessels in which we have measurements. In the above equation $u^j (x_i, t_i; \theta^j) $ and $ A^j(x_i, t_i; \theta^j)$ represent the outputs given the parameters of the neural network for vessel $\# j$. Minimizing this term will encourage the neural networks to fit the available measurements.

\subsection{Loss function: Collocation points}\label{sec:Loss_collocation}
This part of the loss function corresponds to the collocation points, see the blue dots in figure (\ref{fig:domain_1bifurcation}). These are points that we randomly choose inside the arterial domains using a  latin-hypercube sampling strategy \cite{stein1987large}. Over these collocation points, we impose the physical constraints by encouraging the right hand side of equation (\ref{equ:PINNS_residual}) to be equal to zero. The partial derivatives in the residual expression (equation (\ref{equ:PINNS_residual})) can be computed using  automatic differentiation \cite{paszke2017automatic, abadi2016tensorflow}. This objective is imposed to encourage the neural networks to find a particular set of parameters that make their predictions consistent with  the underlying differential equations, which translates to having the minimum residual. By satisfying this condition together with matching the measurements at particular points in section (\ref{sec:Loss_measurement}), we can obtain a model that is capable of inferring the solution at any spatial coordinate of the domain and any time. This collocation loss function takes the following form (take vessel $\# j$ as example):
\begin{align}
    \mathcal{L}_{\textrm{residual}}^j = \frac{1}{N_r^j} \sum_{i=1}^{N_r^j} ( r_A^j(x_i, t_i ; \theta^j))^2 + \frac{1}{N_r^j} \sum_{i=1}^{N_r^j} ( r_u^j(x_i, t_i ; \theta^j))^2 + \frac{1}{N_r^j} \sum_{i=1}^{N_r^j} ( r_p^j(x_i, t_i ; \theta^j))^2, \quad \quad j = 1, \dots, D
\end{align}
where $N_r^j$ represent the number of collocation points in vessel $\# j$, meaning many blue dots we have in figure (\ref{fig:domain_1bifurcation}). Also, $D$ denote the total number of vessels in our arterial network. The terms $r_A^j(x_i, t_i ; \theta^j)$, $r_u^j(x_i, t_i ; \theta^j)$ and $r_p^j(x_i, t_i ; \theta^j)$ represent the residual of area, velocity and pressure defined in equation (\ref{equ:PINNS_residual}), respectively.

\subsection{Loss function: Interfaces}
\label{sec:continuity}
One-dimensional models can be extended to treat splitting and merging arteries by imposing proper boundary conditions \cite{sherwin2003computational}. In this case, one needs to provide  boundary conditions for each artery at the interface points to ensure conservation. In conventional numerical methods (e.g., Discontinuous Galerkin method \cite{cockburn2012discontinuous, sherwin2003one}), the velocity $u$ and the area $A$ can be discontinuous at the interface points (e.g., bifurcations, junctions), so in order to find the values, a Riemann problem has to be solved. This is typically done by employing the characteristic variables of the hyperbolic system of equation (\ref{equ:systemOfEquations}), accounting for a decoupled system of scalar equations, so that the travelling waves can reach the splitting points \cite{sherwin2003one}. The proposed method can work without using information on the characteristics, instead just imposing the momentum and mass conservation equations suffices. To illustrate our workflow, let us consider the case where the artery $\#1$ splits to $\#2$ and $\#3$, see figure (\ref{fig:domain_1bifurcation}). For each vessel $j$, the area, velocity, pressure and density are denoted as $[A^{j}, u^{j}, p^{j}]$, the spatial and temporal variables are denoted as $[x_{i}, t_{i}]$. In reference to figure (\ref{fig:domain_1bifurcation}), this corresponds to the green stars and the $i$ index on how many of these we use to impose the continuity. The boundary conditions at the bifurcation points are derived by the conservation of momentum and mass \cite{sherwin2003computational} for the parent and daughter vessels:
\begin{equation}
\begin{split}
    &A_{1}u_{1} =  A_{2}u_{2} +  A_{3}u_{3},\\
    &p_{1} + \frac{1}{2} \rho u_{1}^2  = p_{2} + \frac{1}{2} \rho u_{2}^2, \\
    &p_{1} + \frac{1}{2} \rho u_{1}^2 = p_{3} + \frac{1}{2} \rho u_{3}^2.
\end{split}
\label{equ:continuity}
\end{equation}

In the case of more than one bifurcations, these conditions are imposed to every splitting point using the appropriate notation, depending on the numbering of the domains. In the case of junctions, instead of bifurcations, or mixed change of geometry, system (equation \ref{equ:continuity}) takes the form of the appropriate condition following the conservation laws. In general, the interface loss has the form (take bifurcation point $\# k$ as example):

\begin{align}
    \mathcal{L}_{\textrm{interfaces}}^k = &\frac{1}{N_b^k} \sum_{i=1}^{N_b^k} ( A_1^k (x_k, t_i; \theta_1^k)u_1^k (x_k, t_i; \theta_1^k) - A_2^k(x_k, t_i; \theta_2^k)u_2^k(x_k, t_i; \theta_2^k) - A_3^k(x_k, t_i; \theta_3^k)u_3^k(x_k, t_i; \theta_3^k))^2 + \\
    &+ \frac{1}{N_b^k} \sum_{i=1}^{N_b^k} ( p_1^k( x_k, t_i;\theta_1^k) + \frac{1}{2} u_1^k(x_k, t_i ;\theta_1^k)^2 - p_2^k( x_k, t_i;\theta_2^k) - \frac{1}{2} u_2^k(x_k, t_i ;\theta_2^k)^2)^2+ \\
    &+ \frac{1}{N_b^k} \sum_{i=1}^{N_b^k} ( p_1^k( x_k, t_i;\theta_1^k) + \frac{1}{2} u_1^k(x_k, t_i ;\theta_1^k)^2 -p_3^k( x_k, t_i;\theta_3^k) - \frac{1}{2} u_3^k(x_k, t_i ;\theta_3^k)^2 )^2 , \quad \quad k=1, \dots, D_I
\end{align}
where the indices $1$, $2$, $3$ in $\mathcal{L}_b^k$ denote the father and daughter vessels, respectively, at each bifurcation. Also, $D_I$ denotes the total number of bifurcation points in our arterial network. $N_b^k$ represent the number of collocation points on the interface boundaries, i.e. how many green stars we have in figure (\ref{fig:domain_1bifurcation}). So, in the above equation if we choose for example $p^1_1$ that means we calculate the pressure at the interface point with index $k=1$ using the network corresponding to the father vessel, in the notion of figure (\ref{fig:domain_1bifurcation}) this would correspond to domain $1$. For the interface, we feed the neural network the inputs $[x_k, t]$, where $x_k$ is the coordinate of the interface point. Minimizing the interface loss encourages the neural network to satisfy the conservation laws at the interface points.

The interface loss function ensures the information of the measurements in one domain can be propagated throughout the neighboring domains. In the original formulation of Sherwin {\it et. al.} \cite{sherwin2003computational} a Riemann problem was solved on the interfaces due to the hyperbolic nature of the wave propagation equations. Here we have experimented with including a constraint ensuring the propagation of characteristic variables at interfaces, but we did not observe any substantial improvement in terms of accuracy and computational efficiency, thus we choose to not include these additional  constraints in the neural networks' loss function.

\subsection{Loss function: Accounting for all constraints}
In the previous sections, we provided details on how to construct the individual terms of the loss function corresponding to the measurements, physical constraints and continuity, that the model should satisfy. In this section, we will show how to combine these individual terms to create the joint form of the loss function, together with some examples to clarify the procedure. The loss function of our physics informed neural networks (PINNs) takes the form:
\begin{align}\label{equ:total_loss}
    \mathcal{L} = \sum_{j=1}^{D_{M}} \mathcal{L}_{\textrm{measurements}}^j + \sum_{j=1}^{D} \mathcal{L}_{\textrm{residual}}^j + \sum_{k=1}^{D_{I}} \mathcal{L}_{\textrm{interfaces}}^k,
\end{align}
according to the definition in previous section, $D$ represent the total number of domains, $D_{M}\subset D$ denotes the indices of domains where we have data and $D_{I}$ correspond to the number of interfaces in the arterial networks. The index $j$ denotes the identity of the domain under consideration and the index $k$ is helping as keep track of the interface points. Index $k$ does not refer to specific neural network, because $A$, $u$ and $p$ on the bifurcation points are calculated through the father and daughter vessels by their respective network. This complete form of the loss function that encourages the model to learn the underlining physics of the problem by using measurements inside the domain, not necessarily at the boundaries e.g., figure (\ref{fig:domain_1bifurcation}), and by respecting the imposed physical constraints. We will present how the loss function is constructed for some simple cases in order for the method to be more clear to the reader.

\begin{figure}
\centering
\includegraphics[width=0.5\textwidth]{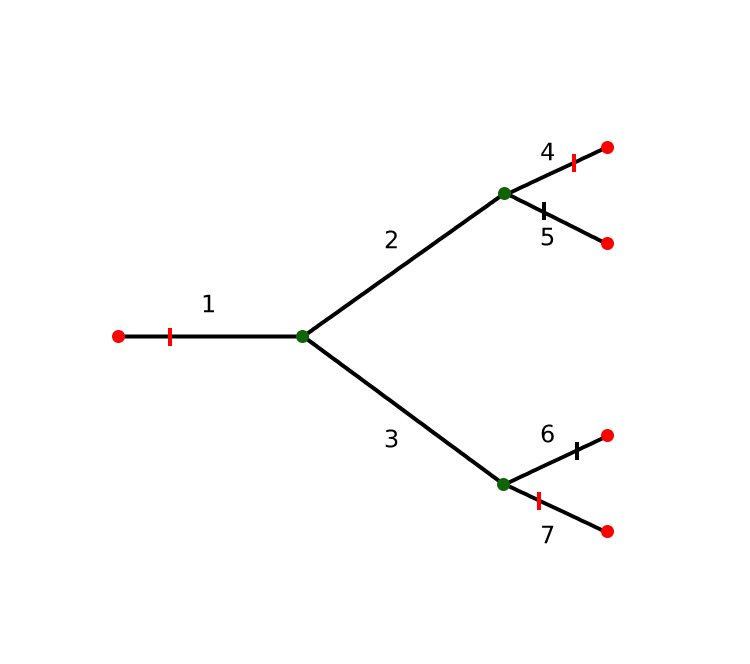}
\caption{{\em Graph of a complex arterial network which involves 7 vessels and 3 bifurcation points:} The red points represent the boundaries of the arterial networks. The gray points represent the bifurcation points where we impose the continuity boundary conditions. The number on each vessels are the corresponding label for these vessels. The vertical lines denote the middle points of the vessels and also the spatial locations for which we have training data, e.g., for the velocity (red) and the wall displacement (black).}
\label{fig:tree_graph}
\end{figure}

For a single artery, the loss function for our method is the following:
\begin{align}
\mathcal{L} = \mathcal{L}_{\textrm{measurements}} +  \mathcal{L}_{\textrm{residual}}.
\end{align}
In this particular geometry there are no bifurcations, so $k=0$, thus the term corresponding to splitting, $\mathcal{L}_{\textrm{interface}}^k$, is neglected. 

For the case that we have three vessels and one bifurcation, see figure (\ref{fig:domain_1bifurcation}), we specify the loss function as:
\begin{align}
\mathcal{L} = \sum_{j=1}^{3} \mathcal{L}_{\textrm{measurements}}^j + \sum_{k=1}^{1} \mathcal{L}_{\textrm{interfaces}}^k + \sum_{j=1}^{3} \mathcal{L}_{\textrm{residual}}^j ,
\end{align}
where the first summation corresponds to the domains where we have measurements for the velocity and cross-sectional area. The second summation encourages the model to satisfy the continuity equation on the bifurcation point. The third summation represents the physics-informed constraints in all three vessels. The second term is written as a summation, for consistency, but in this case, $k=1$, so there is only one component of $ \mathcal{L}^k_{\textrm{interface}}$.

\begin{figure}
\centering
\includegraphics[width=1\textwidth]{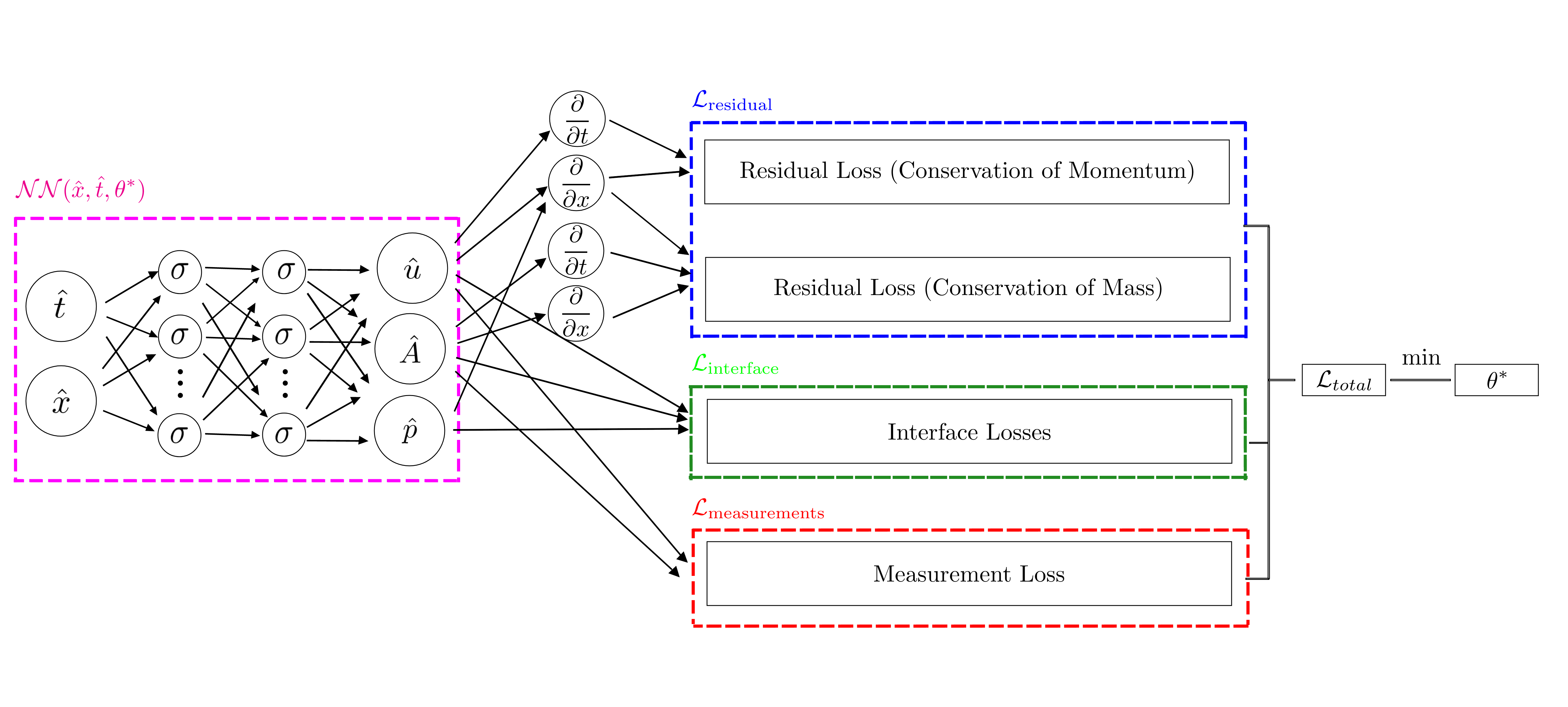}
\caption{{\em A schematic representation of the proposed algorithm.} A diagram describing the process followed in the proposed method is presented. The blue colored box represents the part of the loss function that corresponds to the residual losses, the green colored box represents the part of the loss function that corresponds to the interface losses and the red colored box represents the measurement loss. The different parts construct the complete loss function, by minimizing which the neural network parameters are tuned.}
\label{fig:pinns_arch}
\end{figure}
For a network with 7 vessels and 3 bifurcations, see figure (\ref{fig:tree_graph}), we specify the loss function as:
\begin{align}
\mathcal{L} = \sum_{j=1}^{7} \mathcal{L}_{\textrm{measurements}}^j + \sum_{k=1}^{3} \mathcal{L}_{\textrm{interfaces}}^k + \sum_{j=1}^{7} \mathcal{L}_{\textrm{residual}}^j.
\label{equ:PINN_loss}
\end{align}
Here, the first summation corresponds to the seven vessels where we have measurements of the velocity and area. The second summation correspond to the three bifurcation points. The third summation is representing the physics-informed constraints on seven different arteries. The loss is minimized by encouraging the model to reconstruct the measurements provided at certain points where the measurements are obtained and to respect the constraints imposed by the physics of the problem, e.g., the differential and continuity equations. The loss function in equation (\ref{equ:PINN_loss}) is minimized using stochastic gradient descent algorithm \cite{ruder2016overview} or its modern variants \cite{kingma2014adam, duchi2011adaptive} that are widely used in current deep learning research. 

\subsection{Non-dimensionalization and normalization}\label{sec:scalnorm}
In equation (\ref{equ:systemOfEquations}), the order of magnitude of the different physical quantities, velocity, cross-sectional area and pressure, have a significant relative difference, e.g., $ P\sim 10^{6}$ $ Pa$, $A  \sim 10^{-5}$ $m^2$ and $u \sim 10$ $m/s$, which casts great difficulty on the training of the neural network  \cite{glorot2010understanding}. {\color{black} The significant difference in magnitude of the parameters creates a systematic problem for the training of the physics-informed neural network, as this difference in scales will have a severe impact on the magnitude of the back-propagated gradients that adjust the neural network parameters during training.} To overcome this problem, we employ a  non-dimensionalization and normalization technique with the purpose of scaling the input and the output of the neural networks in a proper scale (e.g., $(\hat{A}, \hat{u}, \hat{p}, x_*, t_*) \sim \mathcal{O}(1)$) and normalizing the spatial and temporal coordinates to have zero mean and unit variance for training the neural networks more efficiently. {\color{black} Although there could be a way to weight the components of the loss function to mitigate the bias casted into the loss function due to this discrepancy across scales, this process would require a lot of guess-work and tuning. On the other hand, the proposed non-dimensionalization strategy achieves the goal of normalizing the variables in a physically justified and intuitive manner that adheres to the requirements of standard neural net initialization strategies (e.g., Xavier initialization) and yields a robust workflow that is free from ad-hoc hacks and guesswork.} For the purpose of non-dimensionalization we introduce characteristic variables, which are commonly used in multi-scale physics modeling  \cite{famiglietti1994multiscale} in order to simplify the equations. For this problem we need a characteristic length $L$ and a characteristic velocity $U$.  
 
We will choose the characteristic length to be the square root of the mean of the equilibrium cross-sectional area of the network vessels. In order to choose the characteristic velocity we make use of the physiological condition that the wave speed in a vessel has to be one order of magnitude larger than the length \cite{sherwin2003one}, given that, in the normalized length case $c = 10$. Thus:

$$L = \sqrt{ \frac{1}{N} \sum_{j=1}^{D} (A^j_0)}  ,\quad\quad\quad U = 10,$$
where $j = 1, ... , D$. At this point we define the quantities:
\begin{equation}\label{equ:non_dimensionalization}
    \hat{u} = \dfrac{u}{U},\quad \hat{A} = \dfrac{A}{A^o},\quad \hat{p} = \dfrac{p}{p_0},\quad x_* = \dfrac{x}{L},\quad t_* = \dfrac{t}{T},
\end{equation}
where $p_0 = \rho U^2$, $T = \dfrac{L}{U}$ and $A^o = L^2$.

Now, all dependent variables take values with $\mathcal{O}(1)$ magnitude. {\color {black} The next step will be performed in accordance to standard deep learning practices for ensuring the robustness of the neural network training via back-propagation. Specifically, according to the seminal work of Glorot and Bengio \cite{glorot2010understanding}, normalizing the input to have zero mean and unit variance mitigates the pathology of vanishing gradients in deep networks. Regarding the choice of activation function, given the relatively shallow networks used in this work, we would expect that the use of a hyperbolic tangent activation should lead to robust results \cite{lecun2012efficient}.} 

Thus, in this step we will apply the above mentioned technique to $x_*$ and $t_*$, in the vessel $\# j$, and get:

\begin{equation}\label{equ:normalization}
    \hat{x}^j = \frac{x_*^j - \mu^j_{x_*}}{\sigma^j_{x_*}},\quad \hat{t} = \frac{t_* - \mu_{t^*}}{\sigma_{t_*}},
\end{equation}
where $\mu^j_{x_*}$, $\mu_{t_*}$ the mean value and $\sigma^j_{x_*}$, $\sigma_{t_*}$ the standard deviation of the spatial and temporal coordinates for the vessel $\# j$, respectively, and $\hat{x}^j$, $\hat{t}$ the scaled inputs.
Using the variables stated above, we derive the updated system of equations (take vessel $\# j$ as example):
\begin{equation}
\begin{split} \label{equ:NormSystem}
       & \frac{1}{\sigma_{t_*}}\frac{\partial \hat{A}^j}{\partial \hat{t}} + \frac{1}{\sigma^j_{x_*}}\hat{A}^j\frac{\partial \hat{u}^j}{\partial \hat{x}^j} + \frac{1}{\sigma^j_{x_*}}\hat{u}^j\frac{\partial \hat{A}^j}{\partial \hat{x}^j} = 0, \\
    & \frac{1}{\sigma_{t_*}}\frac{\partial \hat{u}^j}{\partial \hat{t}} + \frac{1}{\sigma^j_{x_*}} \alpha \hat{u}^j\frac{\partial \hat{u}^j}{\partial\hat{x}^j} + \frac{1}{\sigma^j_{x_*}}\hat{u}^j\frac{\partial (\alpha - 1) \hat{u}^j \hat{A}^j}{\partial\hat{x}^j} + \frac{1}{\sigma^j_{x_*}}\frac{\partial \hat{p}^j}{\partial \hat{x}^j} - \frac{K_R}{L U} \frac{\hat{u}^j}{\hat{A}^j} =0,\\
    & \hat{p}^j = \frac{1}{p_0} (p_{ext} + \beta ( \sqrt{\hat{A}^j A^o} - \sqrt{A_0})), \quad \quad j = 1, \dots, D.
\end{split}
\end{equation}

In this non-dimensional and normalized form, all the variables and inputs are scaled to order $O(1)$. This is what effectively enables the training of our physics-informed neural networks in this complex setting. Likewise, at the predicting stage, we first scale the inputs by the characteristic variables as equation (\ref{equ:non_dimensionalization}), i.e. $x$ by $L$, then normalize them following equation (\ref{equ:normalization}). Finally, we revert the predicted quantities $(\hat{A}^j, \hat{u}^j, \hat{p}^j)$ back to their original form $(A^j, u^j, p^j)$ by multiplying with the characteristic variables i.e. $p = \hat{p} p_0$.

In order to be consistent with the derivation above, we have to follow the same procedure for every condition we impose to the model. In this notion we derive the non-dimensional continuity conditions, by inserting the non-dimensionalizing quantities into the conservation laws. By doing so, we get: 

\begin{align}\label{equ:NormContinuity}
    &\hat{A}_{1}\hat{u}_{1} =  \hat{A}_{2}\hat{u}_{2} +  \hat{A}_{3}\hat{u}_{3},\\
    &\hat{p}_{1} + \frac{1}{2} (\hat{u}_{1})^2  = \hat{p}_{2} + \frac{1}{2} (\hat{u}_{2})^2,\label{equ:NormContinuityp12} \\
    &\hat{p}_{1} + \frac{1}{2} (\hat{u}_{1})^2  = \hat{p}_{3} + \frac{1}{2} (\hat{u}_{3})^2.\label{equ:NormContinuityp13}
\end{align}

The above system of equations must be satisfied at any interface, which is achieved by the second term in the loss function in equation (\ref{equ:total_loss}) that forces the model to respect the conservation laws. Finally, the non-dimensional equations (\ref{equ:NormSystem}), (\ref{equ:NormContinuity}), (\ref{equ:NormContinuityp12}) and (\ref{equ:NormContinuityp13}) define the optimization objectives that are used in equation (\ref{equ:total_loss}) for minimizing the residual at the collocation, bifurcation and measurement points respectively. As mentioned above, we have to multiply the predictions of the network by the scaling parameters $p = \hat{p} p_0$, $u = \hat{u} U$ and $A = \hat{A} A^0$, when doing inference or else we get a scaled version of the solutions.

\subsection{Windkessel model parameter identification as post processing}
\label{sec:windkessel}
{\color{black} A crucial aspect in calibrating computational models of blood flow  is related to the proper choice of boundary conditions.} Due to the excessive computational cost of imaging and performing pulsatile flow simulation across the entire human arterial tree, typically only truncated topologies are considered. This introduces the need of modeling the neglected downstream  dynamics in smaller arteries and arterioles; a task that is typically accomplished using lumped parameter Windkessel models \cite{westerhof2009}. {\color{black} There exist other families of boundary conditions, such as coronary boundary conditions \cite{duanmu2019one} and closed-loop boundary conditions \cite{audebert2015closed}, which are important but not employed in this work}. Windkessel models provide us with a way to account for arterial blood pressure in terms of the compliance of the large elastic arteries and the resistance of the smaller arteries, by making an analogy with an electric circuit. In this setup, the compliance plays the role of a capacitor and the resistance of a resistor and the whole circuit helps to correctly model the influence of the small downstream vessels to the blood circulation. In our case we consider the popular three-element Windkessel model with governing equation (take vessel $\# j$ as example): 
\begin{equation}
\label{Windkessel}
    p^j + R^j C^j \frac{dp^j}{dt} - (R^j + Z^j) Q^j - p_{\textrm{inf}} - R^j C^j Z^j \frac{dQ^j}{dt} = 0, \quad \quad j = 1, \dots D_o,
\end{equation}
where $D_o$ represent the total number of outlets where we want to identify the parameters. In the equation above, $Q^j = A^j u^j$ is the flow rate at the outlet, $p_{\textrm{inf}} = 666.5$ Pa denote the downstream pressure that is constant for all vessels. Also, $Z^j$ represents the characteristic impedance \cite{alastruey2008reduced} that can be easily computed using $Z^j = \dfrac{\rho c_0^j}{A_0^j}$, $C^j$ the total arterial compliance and $R^j$ the systemic vascular resistance. The characteristic impedance is chosen in a way that allows the incoming wave to reach $R^j$ and $C^j$ without being reflected \cite{alastruey2008reduced}. The main challenge here relates to choosing the compliance and resistance parameters $C^j$ and $R^j$ such that physiologically correct results are sought.

Some of the work done in estimating Windkessel model parameters is based on modeling assumptions.  Grinberg {\it et. al.} \cite{grinberg2008outflow} proposed a method of imposing a simple resistance boundary condition using the ratio of flow rates of terminal vessels. They also showed that this method is robust and stable and scales to large arterial systems. The accuracy can be guaranteed by choosing appropriate values for the  resistance at each outlet. Ismail {\it et. al.} \cite{ismail2013adjoint} proposed a method for tuning the parameters of a three-element Windkessel parameters for a coupled 3D/0D system by implementing an inverse analysis based on adjoint methods, using the patient-specific pressure information as a target. Spilker {\it et. al.} \cite{Spilker2010} proposed a multidomain method that gets coupled with a reduced order, a three-element Windkessel, model in which by imposing characteristic information of input pressure, one can tune hemodynamic simulations. The pressure characteristics, such as minimum or maximum pressure, are imposed in the input of the domain. Also Spilker {\it et. al.} \cite{spilker2007morphometry} proposed a procedure for imposing and tuning impedance type boundary conditions by adjusting the length of morphometry produced trees to match the outflow. Yu {\it et. al.} \cite{yu1998estimation} presented an online estimator based on extended Kalman filter that uses sets of measurements to estimate state variables and at the same time recover the model parameters. Pant {\it et. al.} \cite{pant2014methodological} presented an algorithm that iterates between a three dimensional and a zero-dimensional model with known pressure, at specific points, to recover the values of three-elemnent Windkessel model using Unscented Kalman filters. Schiavazzi {\it et. al.} \cite{schiavazzi2017patient} proposed a method for automatically estimating Norwood circulation model parameters and their uncertainty related to the clinical measurements. In this framework an analysis over the confidence of the estimations can be performed using a differential evolution adaptive Metropolis estimation and patient specific measurements on pressure and flow. 

All the aforementioned works require the precise prescription of inflow and boundary conditions and rely on the repeated evaluation of conventional solvers for different combinations of $R^j$ and $C^j$ until  a parametric setting that yields a reasonable match to clinical data is identified. For networks with more than a handful of outlets this defines a tedious and prohibitively costly calibration procedure. In contrast, our proposed methodology can yield a cheap and effective procedure for calibrating Windkessel model parameters without the need of employing conventional simulators. Specifically, the training of the proposed physics-informed neural networks (as described in section (\ref{sec:methods})) does not require the a-priori specification of Windkessel-type outflow boundary conditions. Moreover, once the neural networks are trained on scattered velocity and wall displacement data, they can yield physically consistent predictions for the propagating pressure wave. This information can be then used for identifying the resistance and compliance parameters $C^j$ and $R^j$ at each outlet by solving the differential equation (\ref{Windkessel}) as a post-processing step.


To this end, here we propose a simple method for identifying the parameters $R^j$ and $C^j$ in equation (\ref{Windkessel}) by adaptive grid search given the predicted outflow data $p^j$ and $Q^j$ produced by the physics-informed neural networks representing the flow solution at each outlet. Once the model is trained we can predict $p^j$ and $Q^j$ at every locations in the domain and at any times. Thus we have access to the data as a time series $[p^j_0,\dots, p^j_{N_t -1}]$ and $[Q^j_0,\dots, Q^j_{N_t -1}]$ via the neural network output, where $N_t$ is the length of the time series. We select a periodic part of the data and fit $Q^j$ using Fourier series consisting of fifty modes and then compute the time derivative of $Q^j$ with respect to time $t$. Thus, we are setting up a post processing step that leverages the solution for $Q^j$ and its derivative $\dfrac{dQ^j}{dt}$ at each temporal discretization and use the ODE solver to compute $\Tilde{p^j}$ using some combination of parameters $(R^j, C^j)$. In this case, we compare the difference between $\Tilde{p^j}$ and the pressure predicted by the model. Adaptive mesh refinement of the parameter space can be used to discover the optimal solution for $(R^j, C^j)$ by iteratively exploring new $(R^j, C^j)$ meshes after computing the loss function. We define the loss function as a function of $(R^j, C^j)$ at $N_t$ points $[t_{1},\dots t_{N_t}]$ randomly selected within the interval $t\in[0,T]$:
\begin{equation}\label{equ:RCR_loss}
    \textrm{Loss}(R^j, C^j) = \frac{1}{N_t}\sum_{m = 1}^{N_t}\|\Tilde{p}^j(t_{N_t}) - p^j(t_{N_t})\|^2
\end{equation}

In this case, we compute $\textrm{Loss}(R^j, C^j)$ at the parameter grid and find the set of values for which this loss gets minimized. Then we explore an new interval $R^j_{new} \in [R^j_{optimal} -0.5 R^j_{optimal}, R^j_{optimal} + 0.5R^j_{optimal}]$,  $C^j_{new} \in [C^j_{optimal} -0.5 C^j_{optimal}, C^j_{optimal} + 0.5C^j_{optimal}]$ and refine the mesh in this vicinity. This is done for a number of consecutive times. In our case, we consider the number of consecutive times as 5, in order to reach the parameter set that provides us with the smallest value of loss in equation (\ref{equ:RCR_loss}). The computational time required for discovering the Windkessel parameters is about 10 minutes per outlet using a{ \color{black} laptop Thinkpad P52s (CPU: Intel i7-8650U, RAM: 32 GiB, GPU: NVIDIA Quadro P500 2GiB)}.

\subsection{Computational cost}
{\color{black} The cost of our computational workflow is primarily dominated by the training time of the physics-informed neural networks, which, for the largest geometry considered in this work, summed up to approximately 7 hours on a single NVIDIA Tesla P100 GPU card (starting from a random network initialization). For each arterial topology this can be considered as a one-time offline cost, that could be further reduced by leveraging state-of-the-art GPU hardware and parallel multi-GPU implementations. 
Moreover, transfer learning techniques can be employed to reduce the training cost for subsequent simulations (e.g. different patient data-sets, slightly different flow conditions, parametric studies, etc.), as pre-trained networks can be used to initialize new simulations (instead of using a random initialization). The cost of performing predictions with a trained neural network is also negligible compared to its training cost, as the prediction step merely consists of a few matrix multiplications that can be completed in a fraction of a second on a GPU card. Finally, the 10 min cost of calibrating Windkessel model parameters at each outlet using a trained neural network as a surrogate is small compared to the total training cost, and is trivially parallelizable across multiple outlets. This cost is also reasonable in comparison to established approaches for parameter calibration that typically involve the repeated solution of forward partial differential equations  models in an adjoint optimization loop, or employ filtering techniques that require the availability of invasive pressure measurements.}

\section{Results}\label{sec:results}

The proposed method will be tested in three different cases. In the first case, we consider a prototype arterial network resembling a carotid bifurcation that consists of 3 vessels with 1 bifurcation as shown in figure (\ref{fig:domain_1bifurcation}). In the second case, we consider a complex arterial network resembling an idealized pelvic geometry that consists of 7 vessels and 3 bifurcations as shown in figure (\ref{fig:tree_graph}). For both of these two cases the data used to train the model are synthetic data generated using a conventional Discontinuous Galerkin simulator \cite{sherwin2003one}. More specifically, the data set is created by considering the results of the Discontinuous Galerkin method for velocity and wall displacement at a steady state cardiac cycle. For the first case, we also provide a series of comprehensive systematic studies in order to quantify the accuracy and robustness of the proposed methods.  Finally, we also present results for a real clinical case utilizing 4D flow MRI data collected near the aorta/carotid bifurcation of a healthy human subject. {\com In all of our numerical studies the viscous losses coefficient $K_{R}$ was set to zero and $\alpha$ to one under the assumption that viscous losses play a secondary role in the larger systemic arteries considered in this work.}The proposed algorithms are implemented in Tensorflow v1.10 \cite{abadi2016tensorflow}, and computations were performed in single precision arithmetic on a single NVIDIA Tesla P100 GPU card. { \color{black} The choice of single over double precision arithmetic will be justified in a subsequent section.} The neural network models for the first and the third case was trained for, approximately, 1 hour and 20 minutes and for the second case for approximately 7 hours. All code and data accompanying this manuscript will be made publicly available at \url{https://github.com/PredictiveIntelligenceLab/1DBloodFlowPINNs}.

\begin{table}
\centering
\begin{tabular}{|c|c|c|c|c|c|}
\hline
$\text{Example}$ & $\text{\# layers/ \# neurons per layer}$& $\text{Learning rate}$&  $\text{Batch size}$ &  $\text{Number of iterations}$ \\ 
\hline
$\text{Y-shaped}$ &  7/100 &  $10^{-3}/10^{-4}$ & 1024 & 90,000/40,000 \\
\hline
$\text{Pelvic}$ &  7/100 &  $10^{-3}/10^{-4}$ & 1024 & 280,000/40,000  \\
\hline
$\text{Real noisy data}$ &  7/100 &  $10^{-3}/10^{-4}$ & 1024 & 75,000/35,000   \\
\hline
\end{tabular}
\caption{{\em Neural network parameter choices for different examples.} In the above table the network structure, learning rate, batch size and number of iterations are presented for all the examples provided in the manuscript. For the learning rate and number of iterations parameters the slash character serves as an indication that the learning rate changes from $\eta = 10^{-3}$ to $\eta = 10^{-4}$ after the specified iteration number.}
\label{tab:arch}
\end{table}

\subsection{Flow through a prototype Y-shaped bifurcation}\label{sec:1bifurcation}
\subsubsection{Arterial network topology and observed measurements}
Here we consider a simple arterial network resembling a prototype carotid bifurcation. The network consists of 3 vessels with 1 bifurcation where each vessel has lengths $L_1 = 170.3$mm, $L_2= 7$mm, $L_3= 6.7$mm, respectively, with one bifurcation point at $x_b = 170.3$mm.  Somewhere within these vessels we assume we have access to time-series data for the blood velocity and the wall displacement, as is shown in figure (\ref{fig:domain_1bifurcation}). Specifically, we assume $ N_{u} = N_{A} = 413$ measurements at the inlet and outlets of the arterial network for $t \in [0, 3.3]s$ which represents approximately four cardiac cycles with a period of $T = 0.8s$. We should emphasize that we do not assume any measurements for the pressure at any location in the arterial network. 

\subsubsection{Generation of synthetic training and validation data}

For this example we created a set of synthetic data for the cross sectional area and velocity using an in-house Discontinuous Galerkin (DG) solver \cite{sherwin2003one,perdikaris2014fractional,perdikaris2015effective}. Out of the resulting waveforms we choose 4 cycles from the steady state solution as training data for the neural network model. For the DG simulation, we considered nominal values for the blood density $\rho=1060$ Kg/m$^3$ and the blood viscosity $\nu = 3.5$ mPas, respectively  \cite{fossan2018optimization}. Moreover we provide the DG solver with the Windkessel and arterial wall parameters shown in table (\ref{Tab_Physiological_data_Bif}).

\begin{table*} 
\footnotesize
\centering
\begin{tabular}{cccccccc} \hline\hline
Arterial & Length & Peripheral & Peripheral & $\beta$ & Equilibrium cross- \\ 
segment & (cm) & Resistance & compliance  & & sectional area\\ 
 & & (10$^{10}$ Pa s m$^{-3}$) & (10$^{-10}$ m$^{3}$ Pa$^{-1}$) & (Pa/m) & (m$^2$) \\ \hline 
1 & 17.03 & - & - & 6.97e+07 & 1.36e-05\\ 
2 & 0.7   & 0.5251 & 0.3428 & 5.42e+08 & 1.81e-06 \\ 
3 & 0.67  & 0.2702 & 0.6661 & 6.96e+07 & 1.36e-05\\ 
\hline \hline
\end{tabular}
\caption{{\em Generation of synthetic training and validation data for the prototype Y-shaped bifurcation:} Physiological data used as simulation parameters for the Discontinuous Galerkin method.}
\label{Tab_Physiological_data_Bif}
\end{table*}

\subsubsection{Neural network approximation set-up}

We parametrize the solution of this problem using three neural networks, one for each vessel. Each of these networks consists of 7 hidden layers with 100 neurons per layer, followed by hyperbolic tangent activation function. The neural networks are initialized using Xavier initialization \citep{glorot2010understanding}. This architecture has enough capacity to efficiently capture fine features in the propagating waveforms. A more detailed discussion on the neural network structure can be found in section (\ref{sec:sys_study}) where we present a systematic study that justifies the aforementioned choices in terms of balancing the trade-off of computational cost and predictive accuracy. The neural networks take the scaled inputs $[\hat{x}, \hat{t}]$ and predict the non-dimensional outputs $[\hat{A}, \hat{u}, \hat{p}]$, as discussed in section (\ref{sec:scalnorm}). We are providing each neural network with velocity and cross-sectional area time series for one point at the inlet and the two outlets of the arterial network. Using this data we infer the solution at $N_r = 2,000$, randomly chosen (using Latin hypercude sampling \cite{stein1987large}) collocation points, inside each domain, and $N_b = 1024$ interface points. Moreover we randomly choose batches of $N_{batch} = 1,024$ points as input to the Adam optimizer \cite{kingma2014adam} and learning rate $\eta = 10^{-3}$ for the first $90,000$ iterations. Consequently, we set $\eta = 10^{-4}$ for the following $40,000$ for further minimizing the error and fine-tuning the neural network parameters.

\subsubsection{Comparison of model predictions and reference solution}
\begin{figure}
\centering
\includegraphics[width=\textwidth]{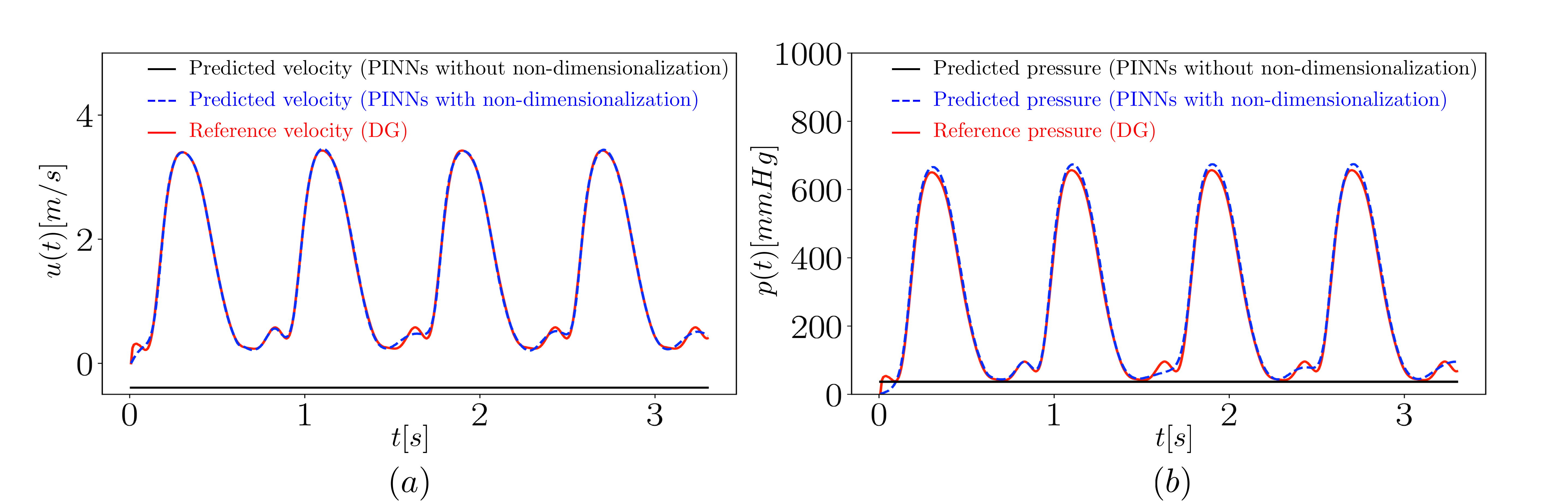}
\caption{{\em Flow through a prototype Y-shaped bifurcation:} (a) Comparison of predicted velocity wave between Discontinuous Galerkin ({\it red}), physics-informed neural networks with non-dimensionalization ({\it blue}) and physics informed neural networks without non-dimensionalization ({\it black}) at the middle point of vessel $\#1$. (b) Comparison of predicted pressure wave between Discontinuous Galerkin ({\it red}), physics-informed neural networks with non-dimensionalization ({\it blue}) and physics-informed neural networks without non-dimensionalization ({\it black}) at the middle point of vessel $\#1$.}
\label{fig:1bifurcation_3vessel}
\end{figure}

We first demonstrate the importance of the non-dimensionalization and normalization of the governing equations, as described in section (\ref{sec:scalnorm}). In order to compute the unknown quantities we assume that we have access to the values of $A(t)$ and $u(t)$ for all times at one spatial point in each vessel, figure (\ref{fig:domain_1bifurcation}). We show in figure (\ref{fig:1bifurcation_3vessel}) that the original methodology introduced in \cite{raissi2017physics,raissi2017physic, raissi2019physics} fails to overcome the difficulty of handling the different scales and thus the numerical prediction collapses. As we mentioned in section (\ref{sec:scalnorm}), this phenomenon occurs due to the significant difference between the orders of magnitude of the model variables, i.e. $\beta \sim 10^{8}$ $Pa/m$, $\rho \sim 10^{3}$ $kg/m^3$, $A \sim 10^{-5}$ $m^2$ and $u \sim 10$ $m/s$, which casts great difficulty on the training of the neural network. We, also, show that the proposed normalization and non-dimensionalization tools significantly improve the model training capabilities and, thus, provide accurate predictions.

\begin{figure}
\centering
\includegraphics[width=\textwidth]{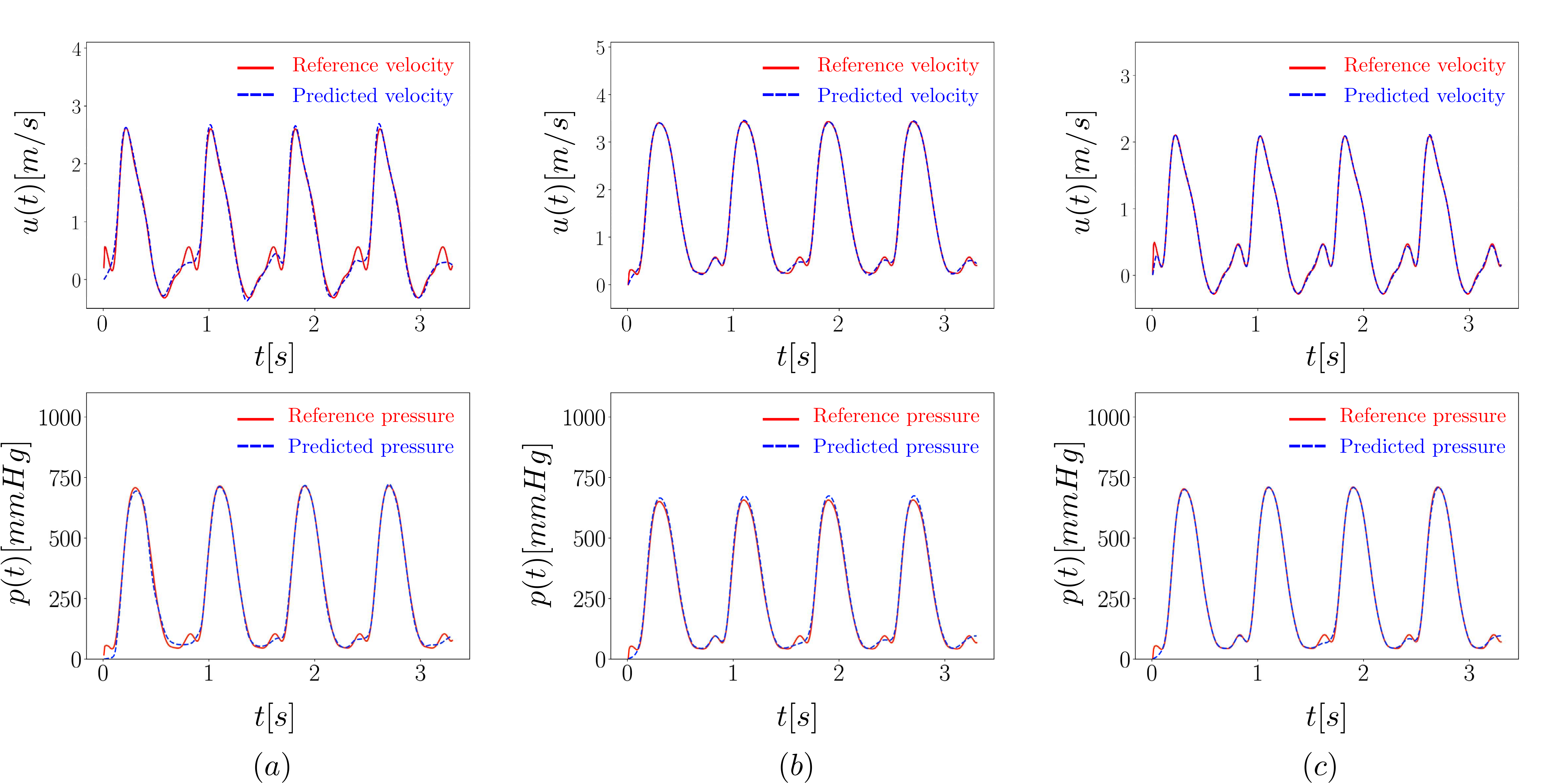}
\caption{{\em Flow through a prototype Y-shaped bifurcation:} Comparison of the physics-informed neural network model predictions and the discontinuous Galerkin reference solution. The numbering of the vessels starts from left to right, with the far left figure corresponding to the father (vessel $\# 1$) and the middle and right one to the daughter (vessels $\#2$ and $\# 3$, respectively), as shown in figure (\ref{fig:domain_1bifurcation}). The top row represents the predicted velocity and the bottom row the predicted pressure for each vessel.}
\label{fig:Compare_1bif}
\end{figure}

 The results produced by our model for the middle points of each domain and randomly selected values of $t$ are presented in figure (\ref{fig:Compare_1bif}). Evidently, there is a good match between the solution acquired from Discontinuous Galerkin solver \cite{cockburn2012discontinuous} and the one predicted by the proposed physic-informed neural networks model. {\color{black} In figures (\ref{fig:1bifurcation_3vessel}) and (\ref{fig:Compare_1bif}) we observe that the secondary peaks are not well captured by the model predictions. We would like to point out that this discrepancy is not related to underlying physiological conditions, as for both computational methodologies the same propagation equations are used, but is mainly attributed to the representation power of the proposed algorithm. The neural network is probably not expressive enough to capture the small details of the wave-form. In our experience, that is a general trait of physics-informed neural network models which often present difficulties in capturing solutions that exhibit multi-scale features. Addressing these shortcomings is still an open area for research and the current best practice for addressing them involves tuning the neural network architecture and/or the optimization algorithm.} The prediction error is further quantified in the systematic studies presented in section (\ref{sec:sys_study}). 

\begin{figure}
\centering
\includegraphics[width=\textwidth]{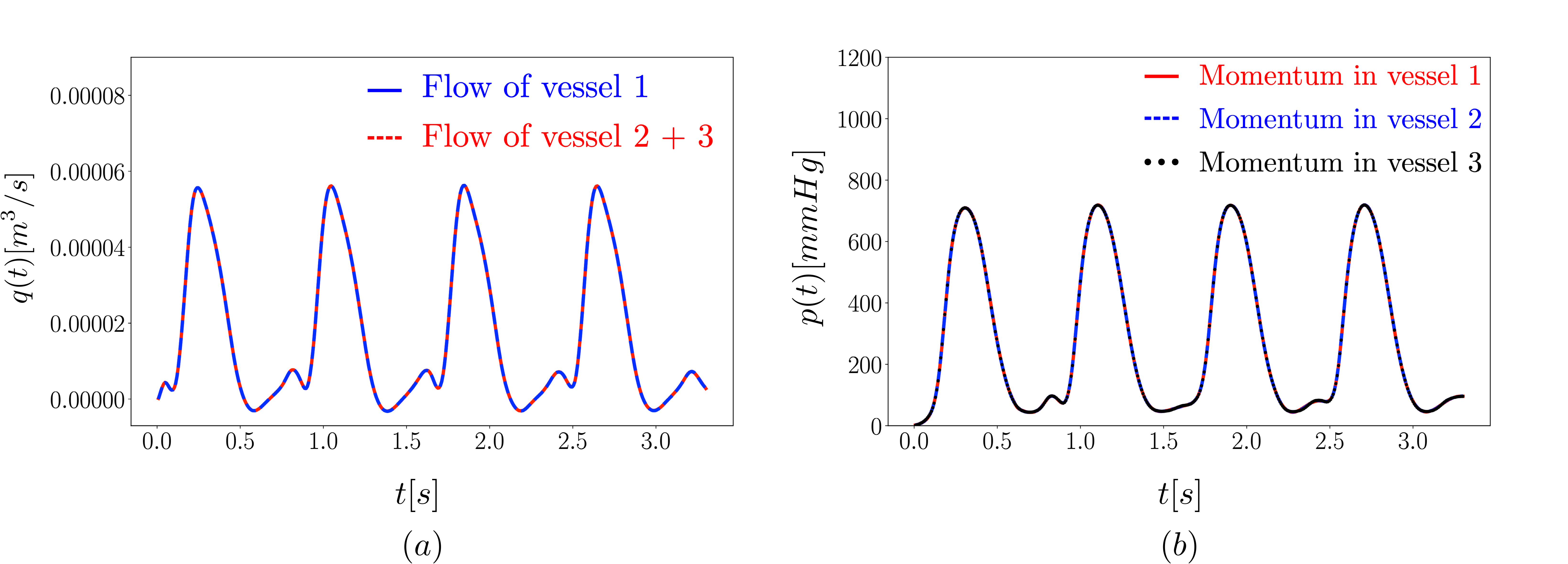}
\caption{{\em Flow through a prototype Y-shaped bifurcation:} Comparison of conserved quantities at the bifurcation point after training the proposed physics-informed neural networks. (a) Conservation of mass (see equation \ref{equ:NormContinuity}). (b) Conservation of momentum (see equations \ref{equ:NormContinuityp12} and \ref{equ:NormContinuityp13}) for the bifurcation point.}
\label{fig:Continuity_equ}
\end{figure}

To highlight the role of the continuity equation described in section (\ref{sec:continuity}), we present a comparison between the left hand side and the right hand side of the continuity equations (\ref{equ:NormContinuity}) and (\ref{equ:NormContinuityp12}),(\ref{equ:NormContinuityp13}) in figure (\ref{fig:Continuity_equ}). We can infer from the figure that the mass in vessel $\#1$ is equal to the total mass in vessel $\#2$ and vessel $\#3$. Moreover, the momentum at the father and daughter vessels at the bifurcation point is equal to each other. Thus, the conservation laws are well preserved in our model.

\subsubsection{Systematic studies}\label{sec:sys_study}
We present a series of comprehensive systematic studies to test the sensitivity of the proposed methods and quantify their robustness with respect to different hyper-parameter settings {\color{black}together with a comparison between single and double precision arithmetic in order to justify our default precision choice.}. All numerical experiments are performed for the canonical Y-shaped bifurcation topology presented in the previous section. To test the sensitivity of the proposed methods with respect to the initialization of the neural networks, we consider a data set comprising of $ N_u = N_A = 413$, $N_r = 2000$, $N_b = 1024$, $N_{batch} = 1024$ training, collocation, interface points and batch size, respectively, and fix the architecture for our neural networks to consist of seven hidden layers containing hundred neurons each followed by a hyperbolic tangent activation function. By doing so, we create an ensemble of fifty cases all starting from a Xavier initialization \citep{glorot2010understanding} for all network weights (with a randomized seed), and a zero initialization for all bias parameters. In tables (\ref{tab:sens_t1}) and (\ref{tab:sens_t1_double}) we report the relative error between the predicted mean solution and the known Discontinuous Galerkin solution \cite{cockburn2012discontinuous} for all 50 randomized trials at point $x=100mm$ in vessel $\#1$ for $2000$ randomly selected temporal values {\color{black}for the single and double precision arithmetic case, respectively.} Evidently, our results are robust with respect to the the neural network initialization as in all cases the stochastic gradient descent \cite{kingma2014adam} training procedure converged to solutions that are close to each other. 
We summarize the result by reporting the mean and the standard deviation of the relative $\mathbb{L}_2$ error {\color{black} defined as}:
\begin{equation}
    \hat{\mathcal{L}}_2 = \frac{\sqrt{\sum_{i=1}^{N_f} (p_{prediction}^i - p_{measurements}^i)}}{\sqrt{\sum_{i=1}^{N_f} (p_{measurements}^i )}},
\end{equation}
for single:
$$\hat{\mathcal{L}_2} \in [\mu_L-\sigma_L, \mu_L+\sigma_L] = [5.29\times 10^{-2} - 2.38\times 10^{-2}, 5.29\times 10^{-2} + 2.38\times 10^{-2}].$$
\begin{table}
\centering
\begin{tabular}{|c|c|c|c|c|c|c|c|c|c|c|}
\hline
\multicolumn{10}{|c|}{Relative $\mathcal{L}_2$ error}\\ 
\hline
3.42e-02& 4.99e-02& 5.65e-02& 3.92e-02& 1.07e-01& 5.52e-02& 5.34e-02& 3.54e-02& 6.41e-02& 4.57e-02 \\
\hline
4.62e-02& 4.04e-02& 2.94e-02& 3.78e-02& 3.48e-02& 1.32e-01& 6.01e-02& 3.00e-02& 4.78e-02& 4.63e-02 \\
\hline
3.68e-02& 4.48e-02& 4.37e-02& 5.81e-02& 1.23e-01& 5.80e-02& 4.06e-02& 2.33e-02& 3.02e-02& 3.96e-02 \\
\hline
5.70e-02& 3.61e-02& 3.85e-02& 3.83e-02& 4.82e-02& 4.85e-02& 5.46e-02& 6.01e-02& 3.84e-02& 7.43e-02 \\
\hline
5.82e-02& 5.09e-02& 4.13e-02& 6.66e-02& 1.31e-01& 7.31e-02& 6.09e-02& 5.59e-02& 3.50e-02& 3.47e-02 \\
\hline
\end{tabular}
\caption{{\em Systematic study on the neural network initialization for single precision arithmetic:} Relative $\mathbb{L}_2$ prediction error for different neural network initialization using different randomized seeds. The prediction errors are obtained by comparing the predicted value of pressure with the reference one in parent vessel $\#1$ at point $x=100mm$ for $2000$ randomly selected temporal values.}
\label{tab:sens_t1}
\end{table}
{\color{black}and double precision arithmetic:}
$$\hat{\mathcal{L}_2} \in [\mu_L-\sigma_L, \mu_L+\sigma_L] = [4.25\times 10^{-2} - 1.99\times 10^{-2},4.25\times 10^{-2} + 1.99\times 10^{-2}].$$
\begin{table}
\centering
\begin{tabular}{|c|c|c|c|c|c|c|c|c|c|c|}
\hline
\multicolumn{10}{|c|}{Relative $\mathcal{L}_2$ error}\\ 
\hline
 4.02e-02& 1.43e-01& 2.21e-02& 3.24e-02& 3.13e-02& 4.77e-02& 2.57e-02& 4.87e-02&  5.66e-02& 2.23e-02\\
 \hline
 2.81e-02& 3.19e-02& 6.22e-02& 3.82e-02& 3.31e-02& 4.87e-02& 3.13e-02& 3.82e-02&  4.62e-02& 2.85e-02\\
 \hline
 3.37e-02& 3.37e-02& 3.26e-02& 3.32e-02& 3.47e-02& 3.62e-02& 3.05e-02& 3.49e-02&  3.40e-02& 2.90e-02\\
 \hline
 3.20e-02& 3.26e-02& 3.58e-02& 2.89e-02& 5.01e-02& 3.99e-02& 3.54e-02& 6.88e-02&  3.35e-02& 5.24e-02\\
 \hline
 2.84e-02& 7.04e-02& 6.08e-02& 6.30e-02& 9.21e-02& 5.36e-02& 5.43e-02& 3.43e-02&  3.85e-02& 3.11e-02\\
 \hline
\end{tabular}
\caption{{\em Systematic study on the neural network initialization for double precision arithmetic:} Relative $\mathbb{L}_2$ prediction error for different neural network initialization using different randomized seeds. The prediction errors are obtained by comparing the predicted value of pressure with the reference one in parent vessel $\#1$ at point $x=100mm$ for $2000$ randomly selected temporal values.}
\label{tab:sens_t1_double}
\end{table} 

{\color{black}It is evident that double precision leads to more accurate results. However, the computational time required for the analysis  is significantly longer (2 hours and 30 minutes instead of 1 hour and 20 minutes, approximately).  Granted that the single precision provides predictions of adequate accuracy, we choose to perform the calculations in single precision to shorten the wall-clock time of the analysis. A potential middle ground to be explored in future work aiming to balance accuracy and computational cost is to employ mixed precision arithmetic: store the network parameters and evaluate its forward pass in single precision, but evaluate the loss functions in double precision.}

Subsequently, we test the sensitivity of our model with respect to the architecture of the neural networks. In this case, we fix the number of noise-free training data as $ N_u = N_A = 413$, $N_r =2000$ and $N_b = 1024$ for each vessel. In all cases, we used a hyperbolic tangent activation function. We initialize our neural network weights using Xavier initialization \cite{glorot2010understanding}. In table (\ref{tab:sens_t3}) we report the relative $\mathbb{L}_2$ prediction error for different fully connected neural network architectures (i.e. different number of layers $N_g$ and number of nodes $N_n$ in each layer). The general trend suggests that as the neural network capacity is increased, we obtain more accurate predictions, which is indicating that our physics-informed constraint on the PDEs residual can effectively regularize the training process and safe-guard against over-fitting. For the rest of the paper we will utilize an architecture defined by seven layers and a hundred neuron per layer for its good performance in balancing the accuracy and computational intensity.  
\begin{table}
\centering
\begin{tabular}{|c|c|c|c|c|}
\hline
\diagbox{$N_g$}{$N_n$} & 20& 50& 100 & 200\\ 
\hline
1 &  6.14e-01 &  8.41e-01 &  7.04e-01 &  1.07e+00 \\
\hline
3 &  4.09e-01 &  3.08e-01 &  3.68e-01 &  2.39e-01 \\
\hline
5 &  5.18e-01 &  2.35e-01 &  1.03e-01 &  2.49e-01 \\
\hline
7 &  4.42e-01 &  3.88e-02 &  3.73e-02 &  3.56e-02 \\
\hline
\end{tabular}
\caption{{\em Systematic study on the neural network architecture:} Relative $\mathbb{L}_2$ prediction error for different fully connected neural network architectures with different number of hidden layers $N_g$, and neurons $N_n$ in each layer. The prediction errors are obtained by comparing the predicted value of pressure with the reference one in parent vessel $\#1$ at point $x=100mm$ for $2000$ randomly selected temporal values.}
\label{tab:sens_t3}
\end{table}

\subsubsection{Windkessel parameter identification by adaptive grid search}

\begin{figure}
\centering
\includegraphics[width=\textwidth]{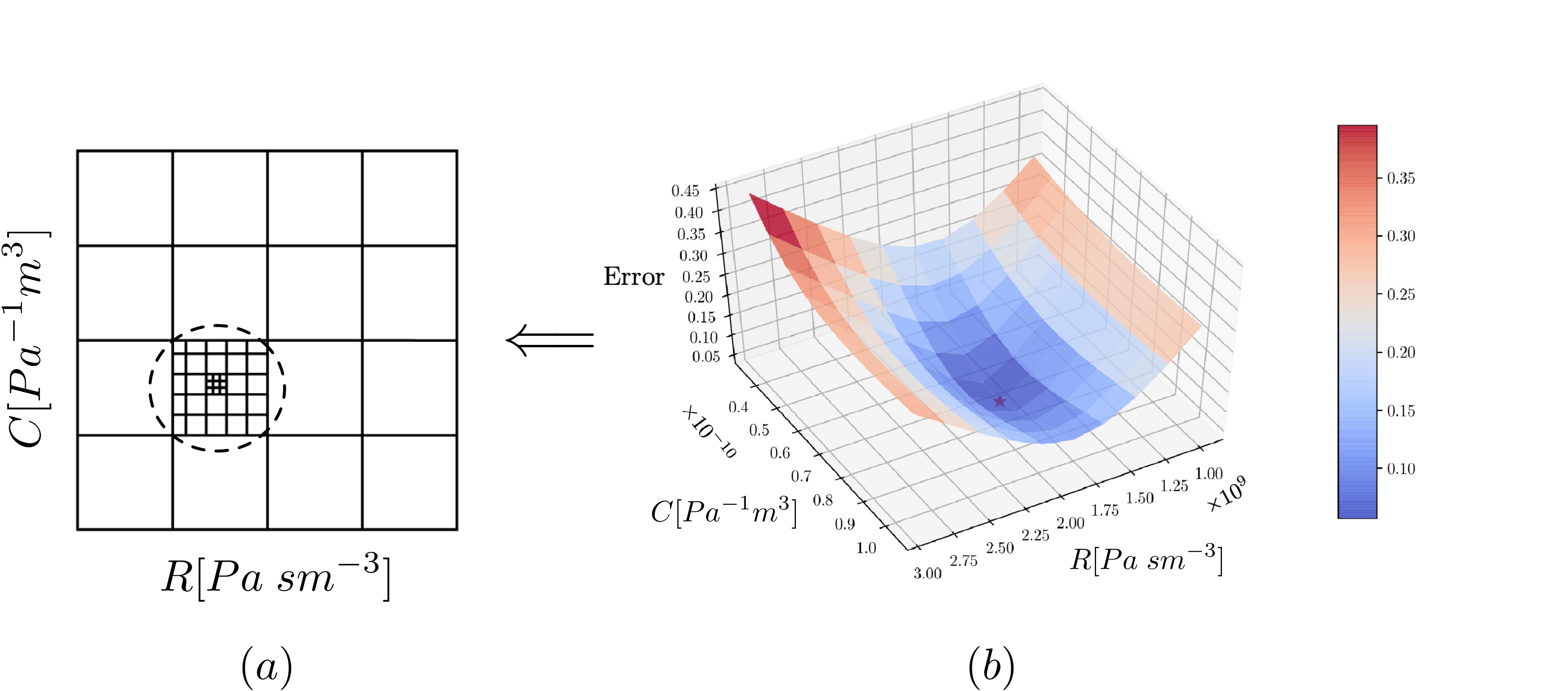}
\caption{{\em RCR identification by adaptive grid search for a Y-shaped bifurcation:} (a) A sketch that illustrates the adaptive grid search. (b) The surface illustrating the values of the loss with respect to parameters $R$ and $C$. The red star denotes the point in the parameter space where the loss function achieves the minimum value.}
\label{fig:RCR_identify}
\end{figure}

As described in section (\ref{sec:windkessel}), our method provides an effective mechanism for calibrating Windkessel model parameters as a simple post-processing step. Specifically, using the predicted flow and pressure at the outflow points, we start by defining a coarse mesh in the space of Windkessel model parameters, and use an ODE solver with inputs $R$, $C$ and the inferred flow to calculate the pressure wave form. After we calculate the pressure waveform for every grid point, we find the one that produces the smallest relative error when compared with the predicted one and refine the mesh near its vicinity. By repeating this procedure for a number of times, in this case five, we acquire a set of $R$ and $C$ that produce the minimum relative error when compared with the target one. In figure (\ref{fig:RCR_identify}) we present the relative error as a function of the parameters $R$ and $C$ for the case of vessel $\#3$ as a visual example of how the method of parameters identification works. For details about the predicted $R$, $C$ and the corresponding relative loss, equation (\ref{equ:RCR_loss}), we refer to table (\ref{tab:RCR_1bifurcation}).

\begin{table}
\centering
\begin{tabular}{|c|c|c|c|}
\hline
\diagbox{$\#$ vessel}{$\textrm{Param}$}& $\textrm{R}_{\textrm{Pred}}$ & $\textrm{C}_{\textrm{Pred}}$ & $\textrm{Relative error of}\;$ P waveforms \\ 
\hline
$\#2$ & 7.58e+08 $Pa s m^{-3}$ &  4.89e-11 $Pa^{-1} m^{3}$ & 3.36e-02 \\
\hline
$\#3$ & 1.84e+09 $Pa s m^{-3}$ &  7.36e-11 $Pa^{-1} m^{3}$ & 3.67e-02 \\
\hline
\end{tabular}
\caption{{\em Windkessel parameter identification for a Y-shaped bifurcation:} Resistance and compliance discovered by adaptive grid search for the three vessels, one bifurcation case and the $\mathcal{L}_2$ relative error between the pressure waveforms predicted by the physics-informed neural networks and the one reconstructed using equation (\ref{Windkessel}) with the identified $R$ and $C$ parameters.}
\label{tab:RCR_1bifurcation}
\end{table}

\subsection{Flow through an idealized pelvic arterial network}

\subsubsection{Network topology and observed measurements}
Our goal here is to demonstrate the effectiveness of the proposed methodology for a slightly more complicated problem, where the available training data lies inside the arterial domain, i.e., no available data is assumed at the inflow or the outflow boundaries. This illustrates a favorable property of our approach versus traditional simulation tools that necessitate the precise prescription of boundary conditions, especially for cases in which measurements are not accessible on the boundaries. To this end, we consider a network consisting of seven arteries with three bifurcations, where every single artery splits into two. This topology resembles an idealized female pelvic geometry, starting from the descending aorta, splitting down to the uterine arteries. We set the vessel lengths to be $L_1 = 10.68mm$, $L_2 = 66.66mm$, $L_3 = 69.94mm$, $L_4 = 147.74mm$, $L_5 = 149.50mm$, $L_6 = 136.42mm$ and $L_7 = 134.38mm$, as shown in figure (\ref{fig:tree_graph}). Each vessel is considered to have its own local coordinate system defined in the interval $x \in [0,L^j]$, where $L^j$ is the length of the respective vessel. The arterial topology includes three bifurcations at the end points of domains $\#1$, $\#2$ and $\#3$, as depicted in figure (\ref{fig:tree_graph}). For this geometry we assume we obtained measurements for the cross-sectional area and velocity over time at the middle points of domains $\#1$,  $\#4$,  $\#5$,  $\#6$ and  $\#7$ and we want to predict the pressure at the boundary points. No information is provided for domains $ \# 2$ and $ \# 3$, and no information about the pressure in any vessel. In order to show that in this case the model can also propagate the information via the boundaries, we aim to infer the pressure at two points in domains $ \# 2$ and $ \# 3$ namely $x_2 = 3mm$ and $x_3 = 6mm$.

\subsubsection{Generation of synthetic training and validation data}

In this example we generated a set of synthetic data for cross-sectional area and velocity using an in-house Discontinous Galerkin solver \cite{sherwin2003one,perdikaris2014fractional,perdikaris2015effective}. Out of the resulting waveforms, we chose 3 steady state cycles as training data. For the DG simulation, we choose the blood density to be equal to $1060$ Kg/m$^3$ and the viscosity $3.5$ mPas  \cite{fossan2018optimization}. Moreover we provide the DG solver with the Windkessel and structural parameters shown in table (\ref{Tab_Physiological_data_Tree}).

\begin{table}
\footnotesize
\centering
\begin{tabular}{cccccccc} \hline\hline
Arterial & Length &  Peripheral & Peripheral & $\beta$ & Equilibrium cross-\\ 
segment & (cm)  & resistance & compliance &  & sectional area\\ 
 & & (10$^{10}$ Pa s m$^{-3}$) & (10$^{-10}$ m$^{3}$ Pa$^{-1}$)& (Pa/m) & (m$^2$) \\ \hline 
1 & 1.0682    & -      & -       & 2.65e+07 & 2.14e-05  \\ 
2 & 6.66638   & -      & -       & 2.60e+07 & 2.21e-05 \\ 
3 & 6.99352   & -      & -       & 2.63e+07 & 2.17e-05  \\ 
4 & 14.7735   & 0.3133 & 16.62   & 2.82e+07 & 1.97e-05\\ 
5 & 14.9503   & 0.1654 & 31.49   & 2.71e+07 & 2.08e-05  \\ 
6 & 13.6421   & 0.1682 & 30.96   & 2.67e+07 & 2.12e-05 \\ 
7 & 13.4384   & 0.2086 & 2.092   & 2.87e+07 & 1.92e-05  \\ 
\hline \hline
\end{tabular}
\caption{{\em Flow through an idealized pelvic arterial network:} Physiological data used as simulation parameters for the Discontinuous Galerkin method.}
\label{Tab_Physiological_data_Tree}
\end{table}

\subsubsection{Neural network approximation set-up}
We parametrize the solution of each artery by a neural network, seven in total, consisting of seven hidden layer and one hundred neurons followed by a hyperbolic tangent activation function, each. The neural network takes the scaled inputs $[\hat{x},\hat{t}]$ and predicts the non-dimensional outputs $[\hat{A},\hat{u},\hat{p}]$ for each domain. For domains $\#2$, $\#3$ we provide only the initial conditions, and no other information on $A$ $u$ and $p$, and train the model to be able to predict these values inside these domain by employing information propagated by the conservation laws on the boundaries. We are randomly, using Latin-hypercube sampling \cite{stein1987large}, select $N_r = 2000$ collocation points in each domain and have $N_A = N_u = 874$ measurements of $u$ and $A$, sampled at the middle points of domains $\#1$, $\#4$, $\#5$, $\#6$ and $\#7$. For domains $\#2$ and $\#3$ we provide $[\hat{x}_{init}, \hat{t}_{init}] $ as inputs, where the $init$ subscript denotes the coordinates and the time of the initial conditions. In this case $\hat{t}_{init}$ is a constant array having zero values and $\hat{x}_{init}$ a constant array comprised of values of equispaced points inside the domain. We choose $N_{batch} = 1024$ randomly selected points at each training step and the same number of interface points $N_b = 1024$ for imposing the continuity conditions. The model is trained employing the Adam optimizer \cite{kingma2014adam} with $\eta = 10^{-3}$ learning rate for the first $280,000$ iterations and $\eta = 10^{-4}$ for the consequent $40,000$. We observe from the results presented in figure (\ref{fig:3bifurcation_reverse}) that there exists some discrepancy between the predicted values for vessels $\#2$ and $\#3$ for which no data is provided. 


\subsubsection{Numerical comparison and Windkessel parameter identification}
\begin{figure}
\centering
\includegraphics[width=\textwidth]{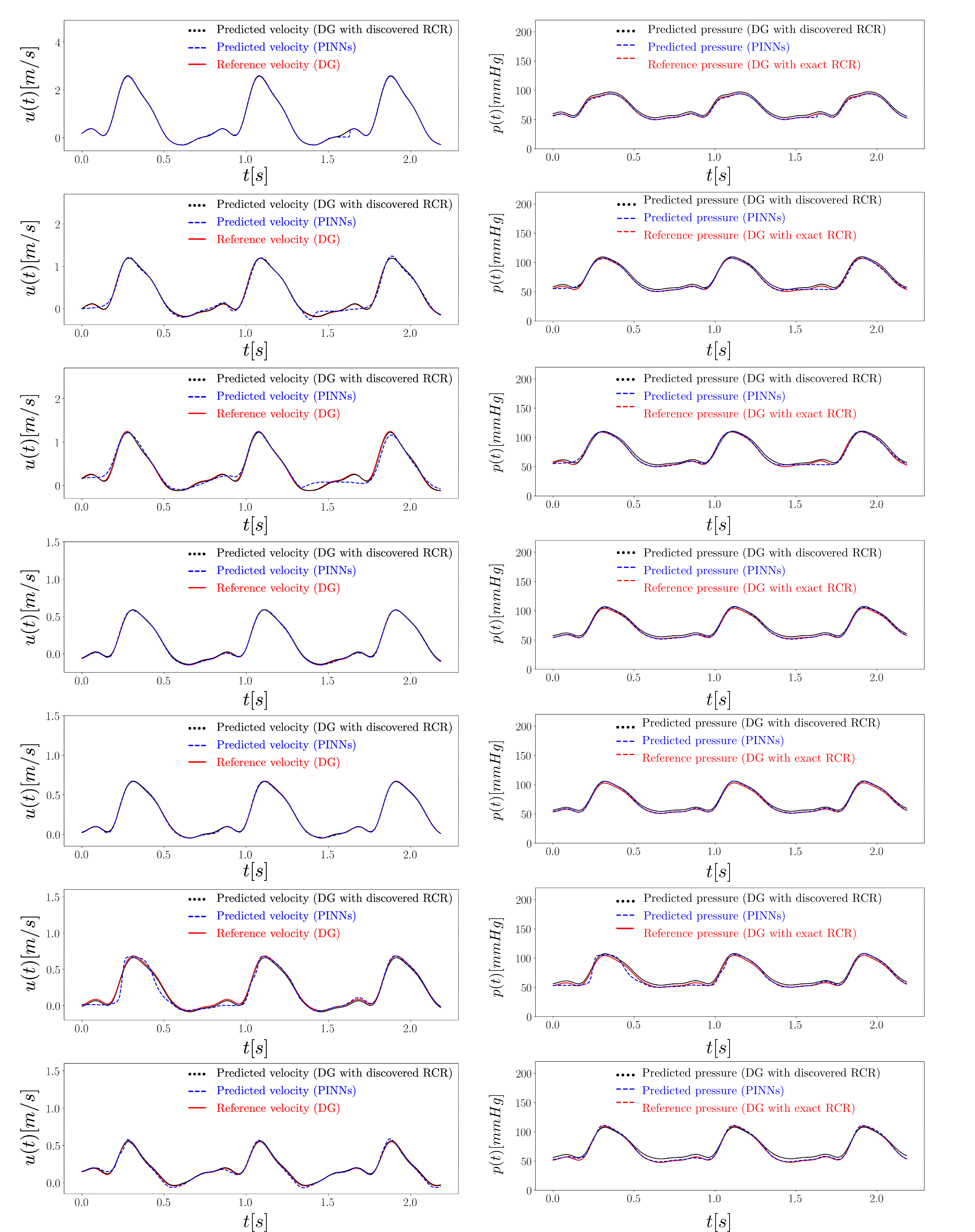}
\caption{{\em Numerical comparison for an idealized pelvic arterial network:} On the left the comparison between the predicted velocity and on the right the predicted pressure of the three methods is presented. The comparison is between our model ({\it blue}), the reference discontinuous Galerkin solution ({\it red}) and the discontinuous Galerkin solution using the discovered Windkessel parameters ({\it black}). On the right, the same case is presented. The vessels are numbered form top to bottom, starting with $\#$1 at the top to $\#7$ at the bottom.}
\label{fig:3bifurcation_reverse}
\end{figure}

In this case, we aim to calibrate a three-element Windkessel model parameters by employing our algorithm to predict the flow at the outflows of the arterial network (red points in figure (\ref{fig:tree_graph})) and then utilize these values, as explained in section (\ref{sec:windkessel}), to solve equation (\ref{Windkessel}). We start by creating a coarse mesh for $R$ and $C$ and solve the ODE on the grid points using the predicted flow. As long as, we calculate the pressure for all grid points, we choose the one for which the relative $\mathcal{L}_2$ error between the ODE solution and the model prediction exhibits the smallest value. Then, we generate an identical mesh around its vicinity. As soon as we calculate the waveforms and their respective relative error for the new grid points, we refine the mesh again. We repeat this process for a number of iterations, five in our case, to acquire the optimal values of resistance and compliance. For details on the predicted $R$, $C$ and the corresponding relative loss we refer the reader to table (\ref{tab:RCR_3bifurcation}). In figure (\ref{fig:3bifurcation_reverse}) the numerical results of our method are presented, in comparison to the solution produced by the discontinuous Galerkin method and the discontinuous Galerkin method using the discovered Windkessel parameters. 

\begin{table}
\centering
\begin{tabular}{|c|c|c|c|}
\hline
\diagbox{$\#$ vessel}{$\textrm{Param}$} & $\textrm{R}_{\textrm{Pred}}$ & $\textrm{C}_{\textrm{Pred}}$ & $\textrm{Relative error of}\;$ P waveforms \\ 
\hline
$\#4$ &  1.36e+09 $Pa s m^{-3}$  &  3.78e-09 $Pa^{-1} m^{3}$ & 7.92e-03  \\
\hline
$\#5$ &  1.64e+09 $Pa s m^{-3}$ &  2.22e-10 $Pa^{-1} m^{3}$& 3.29e-02\\
\hline
$\#6$  & 1.22e+09 $Pa s m^{-3}$ &  2.96e-09 $Pa^{-1} m^{3}$& 8.26e-03\\
\hline
$\#7$ &  2.69e+09 $Pa s m^{-3}$ &  1.84e-09 $Pa^{-1} m^{3}$& 2.56e-02 \\
\hline
\end{tabular}
\caption{{\em Windkessel parameter identification for an idealized pelvic arterial network:} Resistance and compliance discovered by the adaptive grid search for the seven vessels, three bifurcations case and the $\mathcal{L}_2$ relative error between the pressure waveforms of the ordinary differential equation solver using the identified parameters and the model prediction.}
\label{tab:RCR_3bifurcation}
\end{table}




\subsection{Flow through the aorta/carotid bifurcation of a healthy human subject}\label{sec:real_world_data}

For the last example we will test the effectiveness of the proposed method on a realistic clinical case involving measurements acquired in the aorta/carotid bifurcation of a healthy volunteer.

\subsubsection{Geometry and problem setup}
\begin{figure}
\centering
\includegraphics[width=7cm]{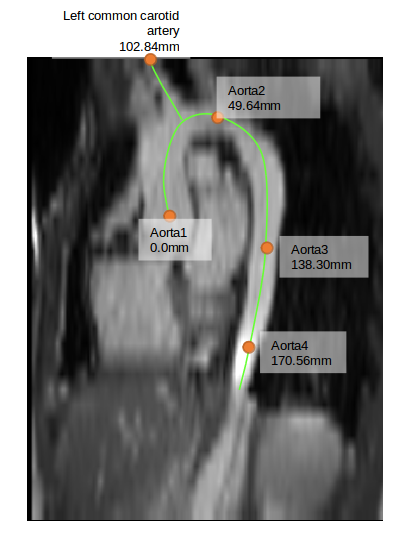}
\caption{{\em Flow through the aorta/carotid bifurcation of a healthy human subject:} Positions of acquired 4D flow MRI measurements in the aorta/carotid bifurcation of a healthy volunteer. Path length measurements are presented in reference to ``Aorta1". Measurements collected at locations [``Aorta1",``Aorta4",``Left common carotid artery"] were used to train the physics-informed neural network model, while measurements at locations [``Aorta2",``Aorta3"] were used to validate its predictions.}
\label{fig:real_world_geometry}
\end{figure}

In this case we consider an arterial geometry consisting of 4 vessels containing 1 bifurcation. An MRI image of this geometry is illustrated in figure (\ref{fig:real_world_geometry}). By utilizing medical image processing and blood flow velocity measuring techniques (see section (\ref{sec:data_acq})), we acquire measurements of area and velocity at the Aorta $\# 1$, Aorta $\# 2$, Aorta $\# 3$, Aorta $\# 4$ and carotid points. Our goal is to leverage the data at the carotid, Aorta $\# 1$ and Aorta $\# 4$ points to predict the area, velocity and pressure anywhere inside the arterial network. More precisely, we will test the predictions of our model against the measurements at Aorta $\# 3$. Furthermore, we would like to utilize the predicted pressure at the carotid point and Aorta $\# 4$ to discover $R$ and $C$ parameters, as discussed in section (\ref{sec:windkessel}). For vessel $\# 3$ we provide the network with just the initial conditions of velocity and area. We adopt a 4 vessels geometry instead of 3, as the equilibrium cross-sectional area is not constant at the vessel ends. Thus, by considering 4 vessels and $A_0$ to be a linear function, we increase the  accuracy of the model. For this case, we choose the neural network architecture and parameters to be the same as in section (\ref{sec:1bifurcation}).

\subsubsection{Clinical data acquisition}
\label{sec:data_acq}
The measurements used in this section were based on acquired MR images from a healthy female volunteer (age=27 years, weight=51 kg, height=160 cm) using a 1.5T Avanto scanner (Siemens Healthcare, Erlangen, Germany). The patient was lying head-first, supine on the table. For the 3-lead electrocardiogram (ECG) monitoring, a neck coil, two body array coils on the chest and abdomen, and a spine array coil were used.The MRI protocol consisted of a Balanced Steady-State Free Precession (bSSFP) localizer of the neck through abdomen, followed by 2D Cine images and prospectively ECG-gated 2D Phase contrast images prescribed at four locations along the aorta and one location in the left common carotid artery, as shown in figure (\ref{fig:real_world_geometry}). Imaging parameters are provided in table (\ref{tab:real_data_details}). The vessel area at each location was quantified by segmenting the 2D Cine images \cite{yushkevich2006user}. The velocity at each location was quantified by manual segmentation of the phase difference images (ImageJ 1.48v;\url{imagej.nih.gov/ij})1.48v;imagej.nih.gov/ij), extracting the mean intensity (mi) inside the
contour at each time frame, and computing the following equation: u = (2048-mi)/2048*venc,
where venc=velocity encoding parameter. The venc values for aorta1, 2, 3, 4, and the carotid
artery were 200, 150, 150, 150, and 150 cm/s, respectively. The arc lengths between consecutive planes were computed by segmentation of the bSSFP localizer images. The centerline extraction, and path length parameterization was acquired by using Seg3D 2.3.0 \url{sci.utah.edu/cibc-software/seg3d.html} \cite{SCI:Seg3D} and VMTK \url{vmtk.org} \cite{antiga2008image}. Finally, we utilized a periodic kernel Gaussian Process regression scheme \cite{rasmussen2004gaussian} to smooth out the input waveforms before we provide them as training data to the physics-informed neural networks. {\color{black} We chose to employ a periodic kernel Gaussian process smoothing over Fourier smoothing for the following reasons. First,  Gaussian process regression naturally returns uncertainty estimates which can be used to assess the goodness of fit and could be subsequently used for uncertainty propagation tasks. Second, the wave-forms we obtain from the clinic may not necessarily be on a regular grid (i.e., the time intervals might not be equi-spaced), and additional interpolation is needed before Fourier smoothing can be concisely applied. Last, in the Fourier case we have to choose the number of modes in such way that the trade-off between approximation quality and noise reduction is optimal. In Gaussian process regression this trade-off between data fit and model complexity is automatically balanced by maximizing the marginal likelihood of the model. }

\begin{table}
\centering
\begin{tabular}{|c|c|c|}
\hline
& 2D Cine & 2D Phase Contrast  \\ 
\hline
Flip angle (degrees) &  55 &  25  \\
\hline
Repetition Time/Echo Time (ms) &  51.74/1.7 &  10.35/6.92 \\
\hline
Bandwidth (Hz/pixel) & 765 &  390 \\
\hline
Voxel size (mm3) &  0.65 x 0.65 x 8 &  0.49 x 0.49 x 5\\
\hline
Temporal Resolution (ms) & 29.4-34.8 &20.7\\
\hline
 $ \# $ Cardiac Phases & 30 (retrospectively-gated) & 37-43 (prospectively-gated)\\
\hline
\end{tabular}
\caption{{\em Flow through the aorta/carotid bifurcation of a healthy human subject:} MRI parameters for 2D Cine images and 2D Phase contrast images used in the clinical data acquisition process.}
\label{tab:real_data_details}
\end{table}

\subsubsection{Numerical results}
\begin{figure}
\centering
\includegraphics[width=\textwidth]{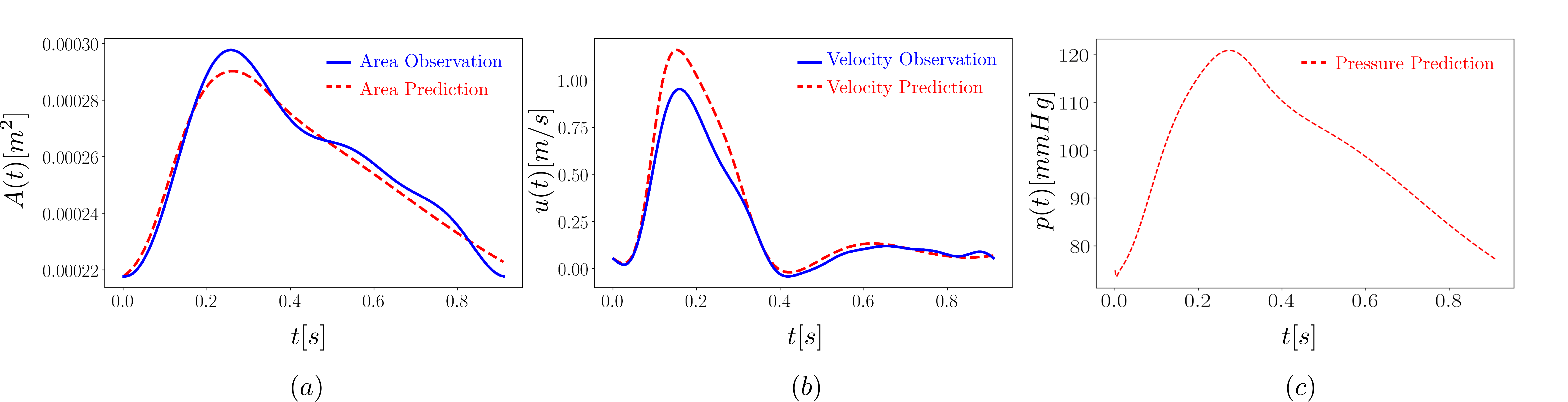}
\caption{{\em Flow through the aorta/carotid bifurcation of a healthy human subject:} (a) Comparison of the neural network model predictions against the clinical cross-sectional area measurements at the test point Aorta $\# 3$. (b) Comparison of neural network model predictions against the clinical cross-sectional velocity measurements at the test point Aorta $\# 3$. (c) Predicted pressure wave at the test point Aorta $\# 3$.}
\label{fig:Real_data_compare}
\end{figure}

\begin{figure}
\centering
\includegraphics[width=12cm]{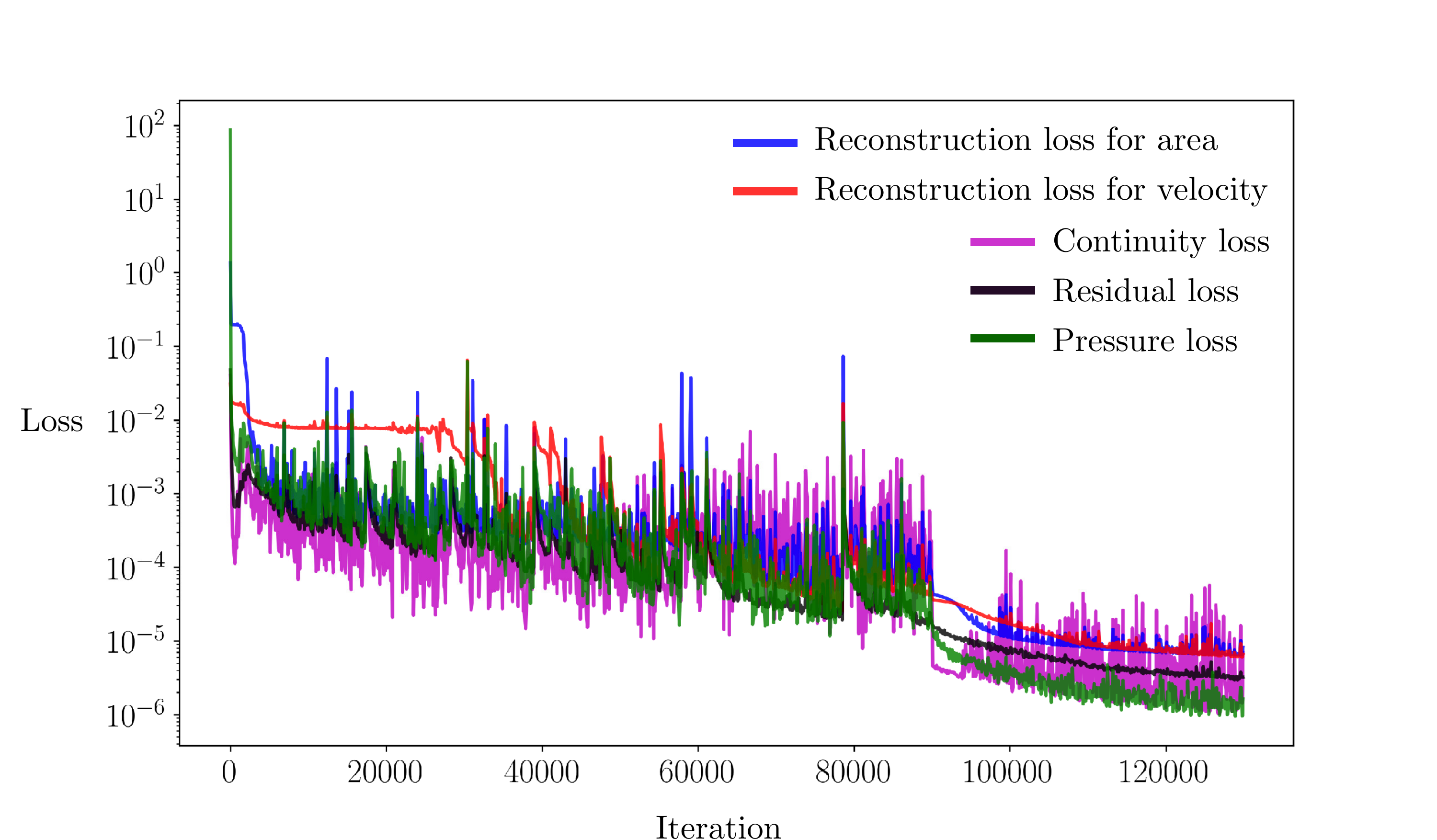}
\caption{{\em Flow through the aorta/carotid bifurcation of a healthy human subject:} Values of the loss functions versus number of stochastic gradient descent iterations during the training of the physics-informed neural networks. The magenta colored line denotes the continuity loss (see equations (\ref{equ:NormContinuity}), (\ref{equ:NormContinuityp12}) and (\ref{equ:NormContinuityp13}) ). The blue colored line denotes the reconstruction loss of the area measurements. The red colored line denotes the reconstruction loss of velocity measurements. The black colored line denotes the residual loss (see equation (\ref{equ:NormSystem})). The green colored line denotes the pressure loss, meaning how close is the pressure predicted by the network with the one produced by inserting the predicted area to the pressure-area relation (see equation (\ref{equ:PINNS_residual})).}
\label{fig:Loss_iteration}
\end{figure}

For this case we will compare the clinically measured data and the model prediction for the Aorta $\# 3$ point. The numerical results are presented in figure (\ref{fig:Real_data_compare}). Comparing the clinical measurements of area and velocity at Aorta $\# 3$ to our model predictions, we show that there is good agreement. Further more, the predicted pressure at this point is within a range consistent with the values reported in literature for a healthy patient (i.e., ranging between 80-120mmHg). We present in figure (\ref{fig:Loss_iteration}) each component of the loss function, equation (\ref{equ:total_loss}), as a functions of the number of training iterations, to exhibit the training efficiency of the method.

As discussed in section (\ref{sec:windkessel}), we can also calibrate the $R$ and $C$ parameters of three-element Windkessel models for each of the outlets using the predicted pressure and flow rate at the carotid and Aorta $\# 4$ points. For more details we refer the reader to table (\ref{tab:RCR_Real}). 

\begin{table}
\centering
\begin{tabular}{|c|c|c|c|}
\hline
\diagbox{$\#$ vessel}{$\textrm{Param}$}& $\textrm{R}_{\textrm{pred}}$ & $\textrm{C}_{\textrm{pred}}$ & $\textrm{Relative error of}\;$ P waveforms \\
\hline
Carotid & 2.09e+09 $Pa s m^{-3}$ &  2.76e-10 $Pa^{-1}m^{3}$ & 3.98e-02 \\
\hline
Aorta $\# 4$ & 1.48e+08 $Pa s m^{-3}$ &  9.23e-09 $Pa^{-1}m^{3}$ & 7.35e-02 \\
\hline
\end{tabular}
\caption{{\em Flow through the aorta/carotid bifurcation of a healthy human subject:} Windkessel model parameters discovered by adaptive grid search, and the $\mathcal{L}_2$ relative error in the pressure wave predicted by the neural networks versus the wave computed by evaluating the 3-element Windkessel model using the identified parameters.}
\label{tab:RCR_Real}
\end{table}

\subsubsection{Discontinuous Galerkin simulation using the identified Windkessel model parameters}

\begin{figure}
\centering
\includegraphics[width=12cm]{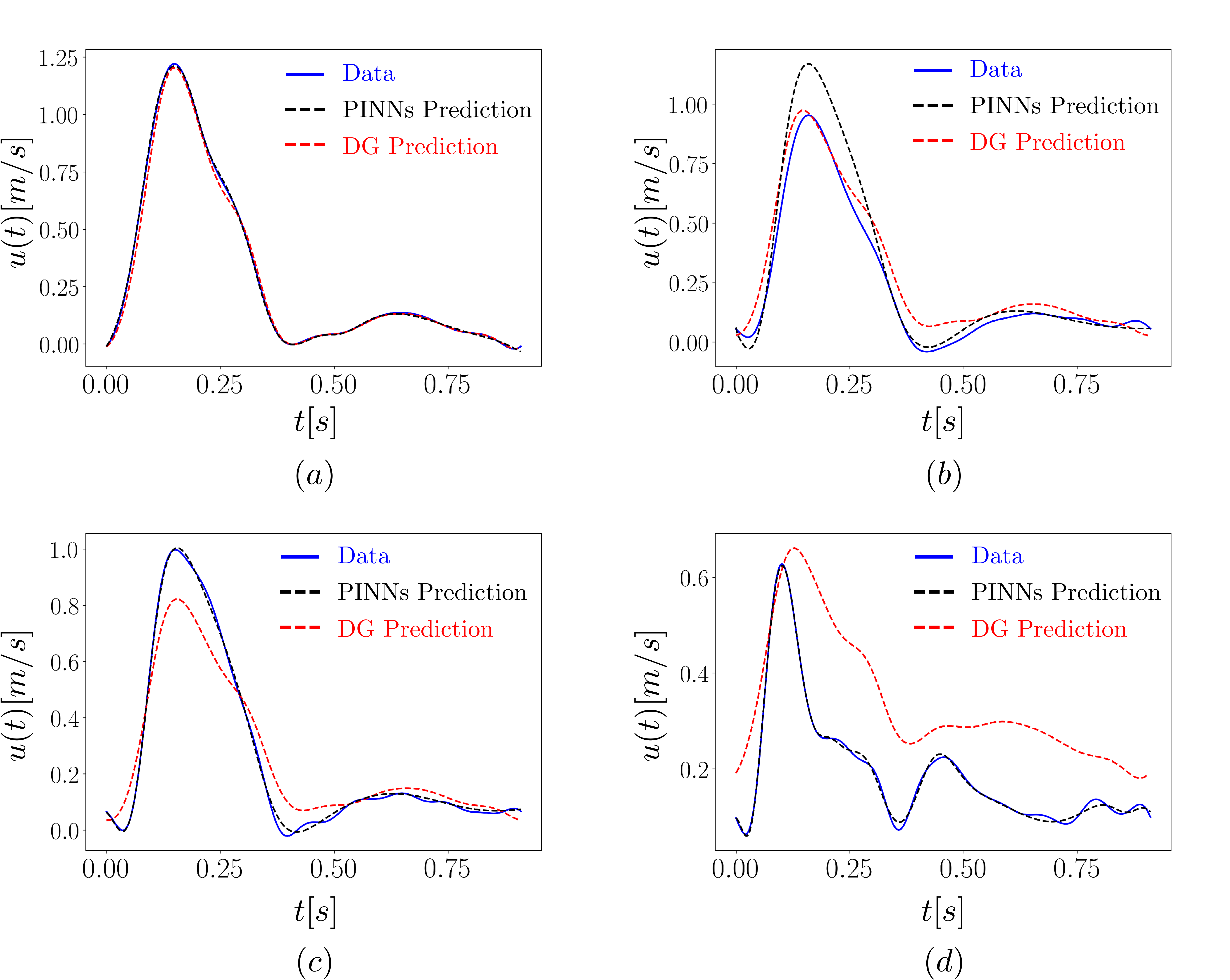}
\caption{{\em Flow through the aorta/carotid bifurcation of a healthy human subject:} Comparison of the clinically acquired waveforms of blood velocity versus the predictions of the proposed physics-informed neural networks, and of a  conventional Discontinuous Galerkin solver with with a three-element Windkessel model outflow condition using the $R$ and $C$ parameters identified by post processing the neural network outputs.  (a) Aorta $\# 1$. (b) Aorta $\# 3$. (c) Aorta $\# 4$. (d) Left Common Carotid.}
\label{fig:RCR_DG}
\end{figure}

Using the discovered three-element Windkessel model parameters we can employ a conventional Discontinuous Galerkin solver to infer the velocity within the arterial network, and compare these with results against the reference measurements and the neural network model predictions. For the DG simulation, we choose the blood density to be equal to $1060$ Kg/m$^3$ and the viscosity $3.5$ mPas  \cite{fossan2018optimization}. Moreover, we provide the DG solver with the Windkessel and structural parameters introduced in table (\ref{Tab_Physiological_data_Real}). The results are presented in figure (\ref{fig:RCR_DG}).

\begin{table}
\footnotesize
\centering
\begin{tabular}{cccccccc} \hline\hline
Arterial & Length & Peripheral & Peripheral & $\beta$ & Equilibrium cross- \\ 
segment & (cm) & Resistance & compliance  & & sectional area\\ 
 & & (10$^{10}$ Pa s m$^{-3}$) & (10$^{-10}$ m$^{3}$ Pa$^{-1}$) & (Pa/m) & (m$^2$) \\ \hline 
1 & 4.964 & - & - & -6.47581e+06*x + 2.47267e+06 & 6.91e-04*x + 2.29e-04 \\ 
2 & 5.32 & 0.2285 & 2.759 &1.37380e+08*x + 2.15121e+06& -4.46e-03*x + 2.64e-04 \\
3 & 8.866 & - & - &7.32369e+06*x + 2.15121e+06& -5.18e-04*x + 2.64e-04 \\ 
4 & 3.2259 & 0.01667 & 92.32 & -8.42729e+06*x + 2.80053e+06 & 7.26e-04*x + 2.18e-04 \\ 
\hline \hline
\end{tabular}
\caption{{\em Flow through the aorta/carotid bifurcation of a healthy human subject:} Physiological data used as simulation parameters for the Discontinuous Galerkin method for the aorta/carotid bifurcation case. Based on the 4D flow MRI measurements, we have assumed a linear tapering across the vessels' length (i.e. the equilibrium  cross-sectional area and the parameter $\beta$ are linear functions of the local spatial coordinate of each artery).}
\label{Tab_Physiological_data_Real}
\end{table}

In section (\ref{sec:simplified_model}) we argued that the presented one dimensional reduced order model \cite{sherwin2003one} constitutes an accurate approximation of the underlying fluid dynamics, but here we do actually observe some discrepancy between the model predictions and the clinically acquired data, see figure (\ref{fig:RCR_DG}) for Aorta $\# 4$ and the carotid outlets. The accuracy of the one-dimensional model has been previously validated by Reymond {\it et. al.} for a both nominal as well as patient-specific geometries \cite{reymond2009validation,reymond2011validationPS}. In both cases the authors considered arterial trees consisting of many systemic arteries including the Left Common Carotid Artery and the Thoracic Aorta. In the case of patient specific validation, they employed an average of 4 measurements of the local diameter at each cross-section and also averaged over 5-15 cardiac cycles to obtain a smoothed flow waveform. Moreover, for modeling the pulse wave propagation they utilized a form of incompressible Navier-Stokes equations that takes into consideration the wall viscous effects \cite{reymond2009validation}, as well as a non-linear viscoelastic pressure relation for which they tuned the viscoelastic parameters based on values appearing in the literature. Even in this more detailed case, where more complex models that can better capture the complexity of the real world are utilized, there existed some discrepancy between the predicted values and the measured ones \cite{reymond2009validation,reymond2011validationPS}. In our case, we do not pre-process the data in such a detailed manner, nor we employ such an expressive model that requires a large amount of information related to geometry, elastic properties etc. It can be seen from figure (\ref{fig:RCR_DG}) that the Discontinuous Galerkin simulation that utilizes the Windkessel parameters discovered by the procedure described in section (\ref{sec:windkessel}), can capture the magnitude and the wave peak timing in a favorable manner. For the case of the Left Common Carotid Artery we detect a larger amount of discrepancy and a time shift, but this maybe attributed to back-propagating elastic waves that the model in this form can not capture. Overall, considering the simplicity of the one-dimensional pulsatile flow model and the amount of provided information, it still produces favorable results. The prediction of the neural network on the other hand is very accurate at the points that measurements are provided, but slightly over-estimates the velocity at the point that the model is not trained. This discrepancy can be attributed to the simplicity of the underlying physics-informed constraints, as well as the noise corruption in the velocity and cross-sectional area measurements used to train the neural network model. Overall, for the limited amount of measurements and the noise level they contain, the performance of the deep learning model still has margins for improvement. 

\section{Discussion}
\label{sec:discussion}

Advances in physics-informed machine learning provide the connecting link for integrating theoretical models and real-world   data. This work is just one example of this general paradigm applied to the modeling and simulation of cardiovascular flows. Here, we have demonstrated how one-dimensional models of blood flow model can by seamlessly synthesized with clinical data to construct deep neural networks that can predict quantities which cannot be reliably measured in a non-invasive manner (e.g., blood pressure), by complementing the governing laws of fluid flow in compliant arteries with scattered measurements obtained by  medical imaging techniques. To facilitate the efficient training of such physics-informed networks, we put forth proper non-dimensionalization and normalization techniques, as well as formulated a composite training objective that enables the consistent propagation of flow information across a networks of systemic arteries. The proposed methodology yielded favorable results across a collection of synthetic and realistic examples, and showed good agreement with state-of-the-art numerical solvers using Discontinuous Galerkin discretizations. Moreover, we showed how a simple evaluation of the trained neural networks can provide a low-cost post-processing procedure for calibrating Windkessel model parameters for cases where traditional simulations are sought.

{\com

In comparison with conventional pure physics-based computational models, the advantages of our data-driven approach can be mainly attributed to the fact that they by-pass the tyranny of mesh generation, and they do not require the precise prescription of initial or boundary conditions, or even constitutive laws \cite{raissi2019physics, tartakovsky2018learning}. One can view the proposed techniques as a tool for PDE-constrained filtering of scattered noisy data in which the underlying physics can be only partially known, as opposed to interpreting these algorithms as numerical PDE solvers in the classical sense. We argue that this can potentially bring additional flexibility and eventually reduce the time required to obtain reliable predictions in a realistic clinical setting.

Deploying the proposed tool in the clinic would necessitate two main directions for improving our present model. First, given measurements for a new patient we need to reduce the problem setup time (i.e., geometry and 4D flow MRI data extraction), as well as the training time of the physics-informed neural network, from several hours to a few minutes in order to make our predictions practically useful to a clinician. The latter can be potentially achieved by leveraging ideas from transfer learning, in which pre-trained models can be used as an initial starting point for training when data for a new patient arrives. In such cases, the neural networks need not be trained from scratch (i.e., from a random initialization), but only be fine-tuned to adapt to the new data. The second direction of improvement is related to enhancing the accuracy and the robustness of the neural network predictions, as well as to endowing those predictions will reliable uncertainty estimates. This would include designing more stable and robust neural network architectures, further understanding the approximation power of physics-informed neural networks, and adopting probabilistic formulations that can enable uncertainty quantification. In all cases, extensive validation studies will be required to assess the potential of physics-informed neural networks as a reliable clinical tool.}

Although the direct incorporation of clinical measurements is one of the main strengths of the proposed methodology, but also one of its weaknesses. Such measurements have often very coarse resolution and may be heavily corrupted by noise. This setting introduces challenges in encouraging the neural networks to simultaneously fit the data and satisfy the underlying conservation laws, as these requirements may contradict each other in the presence of noise. Although, this did not pose a significant challenge in the reported clinical case it may become an issue for studying smaller arterial networks for which the resolution of 4D flow MRI techniques may not suffice to produce readings with low signal-to-noise ratio. {\color{black} Moreover, the use of a mean square error loss tacitly assumes a Gaussian likelihood model for the data corresponding to an isotropic Gaussian noise model which may not be adequate to reflect the noise statistics in the real data signal which are likely non-Gaussian and heteroscedastic. Therefore, a use of a more robust likelihood/loss function or a probabilistic formulation similar to the one recently put forth in \cite{yang2018adversarial} could be useful.  An alternative approach would be implement a physics based de-noising algorithm similar to the one put forth by \textit{Rudy et. al.} \cite{rudy2018deep} in order to de-noise the measured data such that they satisfy the Navier-Stokes equation in a more accurate manner. We could also implement a more detailed form of the Navier-Stokes equation, taking in to account more detailed effects that better reflect the underlying physics (e.g., non-Newtonian rheology, wall visco-elasticity, etc.). On the data front, we could get more samples from the same patient and more patients and calibrate the inputs to have  smoother measurements.

Finally, we should emphasize the need for scalable and reliable methods for uncertainty quantification, especially for over-parametrized deep learning models. Clinical applications like the one studied in this work often involve scarce data and high-consequence decisions, therefore posing the need for robust prediction models. 
Specific to this work, the most straightforward way to quantify uncertainty in the model predictions is to consider ensemble averaging of bootstrapped neural networks \cite{lakshminarayanan2017simple, osband2018randomized}. Alternatively, Bayesian approaches based on Markov Chain Monte Carlo sampling \cite{gelman2013bayesian} or generative models can be employed \cite{yang2018physics}. As examples, we refer the reader to the recent works of \cite{zhu2019physics, yang2018adversarial}, where an example of employing probabilistic neural networks to perform uncertainty quantification for PDE-governed systems is provided. Despite some progress, these works should be considered as first steps in the direction of incorporating uncertainty quantification in deep learning. This is an important aspect in data-driven modeling pipelines (especially in absence of theoretical bounds and guarantees), and defines an open area for future research.}


\section*{Acknowledgments}
The authors would like to thank Veronica Aramendia-Vidaurreta and Brianna Moon for assisting in the acquisition of  carotid/aorta MRI data. G.K., Y.Y. and P.P. would like to acknowledge support from the US Department of Energy under the Advanced Scientific Computing Research program (grant DE-SC0019116), the Defense Advanced Research Projects Agency under the Physics of Artificial Intelligence program (grant HR00111890034). E.H. received support from the Eunice Kennedy Shriver National Institute of Child Health and Human Development (U01-HD087180), National Institute of Biomedical Imaging and Bio-engineering (T32-EB009384) and National Science Foundation (DGE-1321851).


\bibliographystyle{elsarticle-harv}
\bibliography{main}



\end{document}